\documentclass{article}
\usepackage{spconf,amsmath,graphicx}

\usepackage{epsfig}
\usepackage{amssymb}
\usepackage{color}
\usepackage{url}
\usepackage{subcaption}
\usepackage{booktabs}
\usepackage{enumitem}
\usepackage{hyperref}
\usepackage{pdfpages}
\usepackage{bookmark}
\newcommand{\yueyu}[1]{{}}
\title{Standard compliant video coding using low complexity, switchable neural wrappers}

\name{Yueyu Hu$^{\ast \dag}$, Chenhao Zhang$^{\ast \dag}$, Onur G. Guleryuz$^{\ddag}$,  Debargha Mukherjee$^{\ddag}$, Yao Wang$^{\dag}$}
\address{
$^{\dag}$New York University, Dept. Electrical and Computer Engineering, Brooklyn, NY, 11201, USA \\
$^{\ddag}$Google LLC, 1600 Amphitheatre Parkway, Mountain View, CA, 94043, USA} 

\begin{document}

\maketitle
\thispagestyle{empty}

\begin{abstract}

The proliferation of high resolution videos posts great storage and bandwidth pressure on cloud video services, driving the development of next-generation video codecs. Despite great progress made in neural video coding, existing approaches are still far from economical deployment considering the complexity and rate-distortion performance tradeoff. To clear the roadblocks for neural video coding, in this paper we propose a new framework featuring standard compatibility, high performance, and low decoding complexity. We employ a set of jointly optimized neural pre- and post-processors, wrapping a standard video codec, to encode videos at different resolutions. The rate-distorion optimal downsampling ratio is signaled to the decoder at the per-sequence level for each target rate. We design a low complexity neural post-processor architecture that can handle different upsampling ratios.  The change of resolution exploits the spatial redundancy in high-resolution videos, while the neural wrapper further achieves rate-distortion performance improvement through end-to-end optimization with a codec proxy. Our light-weight post-processor architecture has a complexity of 516 MACs~/~pixel, and achieves 9.3\% BD-Rate reduction over VVC on the UVG dataset, and 6.4\% on AOM CTC Class A1. Our approach has the potential to further advance the performance of the latest video coding standards using neural processing with minimal added complexity.

\end{abstract}


\section{Introduction}

\renewcommand{\thefootnote}{\fnsymbol{footnote}}
\footnotetext[1]{Equal contribution.}

Storage and transmission of the ever increasing amount of high-resolution videos has been an emerging challenge for video-on-demand services like Netflix and YouTube. Although high complexity is typically acceptable for cloud encoding for such applications, a strict constraint in decoding complexity is necessary to guarantee seamless video playback at the client side.

To improve the rate-distortion (R-D) performance for high-resolution video coding, apart from the continuous endeavor to improve video coding tools in the hybrid codec~\cite{bross2021overview,han2021technical},  there has been an emerging trend to  introducing neural networks~\cite{liu2020deep,ma2019image,li2022hybrid} into or replacing part of the coding loop. 
Video coding standard expert groups including MPEG, AVS and AOM all have been exploring to standardize neural tools or coding schemes, towards a significant improvement in R-D performance.

Despite the promising rate-distortion improvements neural networks can offer, the huge complexity of neural network computations has been a major roadblock. One thread of works focus on using neural networks to improve certain components in an established hybrid codec. Among these works, learned inter- and intra- prediction incur excessive complexity~\cite{lei2022deep,jin2021deep}. Likewise, convolutional neural filters serving as in-loop deblocking filters, 
require large number of parameters to adapt to various contents. In one example~\cite{liu2022efficient}, the light-weight neural in-loop filter still complicates the decoding by a factor of more than 3 times. The other thread of works adopt end-to-end learned frameworks and have shown more promising R-D performance growth. However, to outperform VVC~\cite{bross2021overview} in R-D performance, even the state-of-the-art efficient method~\cite{wang2023evc} still takes 309K MACs / pixel for decoding, and can only support real-time decoding of 1080p videos on a high-end NVIDIA A100 GPU, making it unpractical for real-world applications. In short, both hybrid neural tools and end-to-end learned neural video compression methods are still far from  practical deployment on consumer-grade decoding platforms.

To this end, we propose a standard compatible neural video coding scheme that boosts video coding performance with low decoding complexity. 
The core idea is to reduce the original high-resolution videos into a lower resolution, which is then coded by a standard hybrid codec.  An efficient neural post-processor is finally used to  upsample the decoded video to the original resolution. In addition to using a linear downsampling filter, we allow the encoder to optionally utilize a neural pre-processor jointly trained with the post-processor. Note that some videos have mostly smooth texture and can be downsampled without losing critical information, while others may contain high frequency details that are hard to recover from a downsampled version by the post-processor. Recognizing that the best down-sampling ratio is content- and rate-dependent, we propose for the encoder to try different downsampling factors and choose the one that leads to the best quality given a particular rate. The best down-sampling factor is signaled as part of the side information at the sequence level.   

Our neural post-processor is designed for low complexity at 516 MAC~/~pixel. With the proposed design, on the UVG dataset we demonstrate an overall 22.6\% BD-rate reduction over HEVC, 9.3\% over the latest VVC standard. Evaluation with AOM CTC Class A1 sequences shows -15.7\% BD-rate over HEVC, and -6.4\% over VVC. The proposed neural post-processor is capable of decoding a 1080p video frame within only 7.7 ms on a mid-range NVIDIA RTX 3060 GPU, and 30.7 ms for a 4K frame. It can be deployed on current desktop computers and potentially TV boxes and mobile devices with neural hardware accelerators. Our contributions are summarized as follows,
\begin{itemize}[leftmargin=*]
    \item We propose a framework for neural network assisted video coding with switchable neural wrappers, which can be  adopted by the next-generation video coding standard to boost R-D performance with low decoding complexity.
    \item We propose an efficient neural network architecture for the post-processor, delivering significant BD-rate reduction with only 516 MACs / pixel complexity.
    \item We show that a jointly trained neural pre-processor can further improve R-D performance without increasing the decoding complexity.
\end{itemize}

\section{Related Works}
\subsection{Neural Processing for Video Coding}
Extensive efforts have been made to use neural network as image processors to improve the compression performance of video codecs~\cite{liu2020deep}. The field of research includes post-processing and in-loop filters~\cite{liu2022qa,ding2021patch,huang2021one,ma2020mfrnet,klopp2021online}, pre-editing and smoothing~\cite{talebi2021better,klopp2021exploit}, and joint pre- and post-processing~\cite{isik2023sandwiched}. Among these works, the most relevant ones to this work are the online-learned upsampler~\cite{klopp2021online} and the sandwiched codec~\cite{guleryuz2022sandwiched}. The work in \cite{klopp2021online} proposes to first spatially downsample a video using a standard linear filter, code the down-sampled video using HEVC, and then upsample using an online trained video-specific post-processor. Since the post-processor is specifically optimized for each individual video segment of up to 120 frames and the post-processor is light weight, the upsampler is very effective at recovering high frequency information with limited number of parameters. However, such a scheme requires training a neural network at the time of encoding each video segment, for each quantizaion parameter (qp). Meanwhile, \cite{guleryuz2022sandwiched} presents the \textit{sandwiched codec} that jointly optimizes pre- and processor for image compression. It shows that it is practical and beneficial to compress a processed image at a lower resolution and upsample for reconstruction with a neural network. However, they used a complicated post-processing neural network architecture and only considered image compression. In this work, we are inspired by the idea of the sandwiched codec while we aim to make the decoding efficient enough for practical use. We show that by including a jointly trained pre-processor and allowing switching between different downsampling ratios, we are able to achieve improvements in R-D performance over a large rate range while maintaining a low decoding complexity.

\subsection{End-to-End Learned Video Compression}
Since the success of end-to-end learned image compression~\cite{balle2018variational,hu2021learning}, there has been tremendous advances in end-to-end learned video compression~\cite{li2021deep,mentzer2022vct,xiang2022mimt}. Because this type of methods conduct per-pixel motion compensation and jointly optimize the entropy encoder, the latest scheme has been shown to surpass VVC in R-D performance \cite{xiang2022mimt}. However, they are usually too complex for immediate deployment, especially on mobile devices. In this work, we seek to build an efficient decoder based on existing state-of-the-art hybrid video codecs, and potentially exploiting the standard-compliant hardware decoder and neural accelerator for practical implementation.

\section{Proposed Method}

\subsection{Coding Framework}

\begin{figure*}[t]
    \centering
    \begin{subfigure}{0.76\linewidth}
    \includegraphics[width=0.95\linewidth]{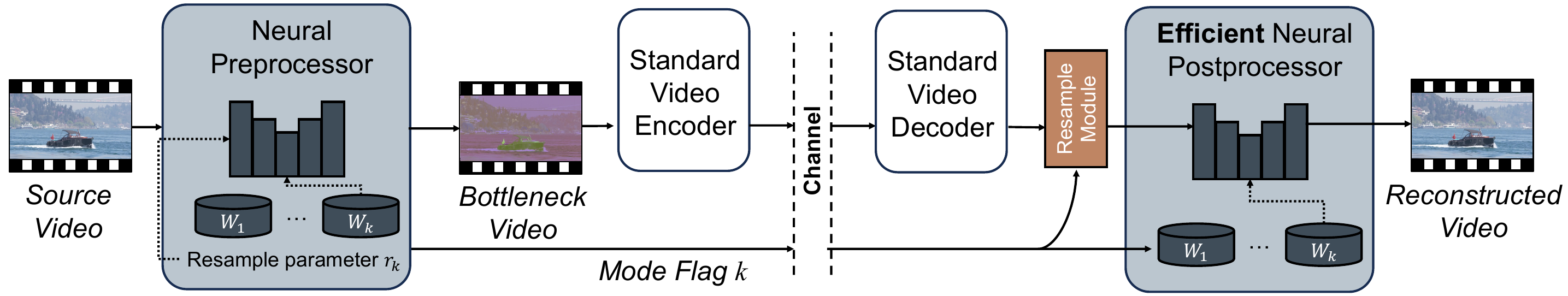}
        \caption{Coding Framework}
        \label{fig:arch}
    \end{subfigure}
    \begin{subfigure}{0.23\linewidth}
        \includegraphics[width=0.95\linewidth]{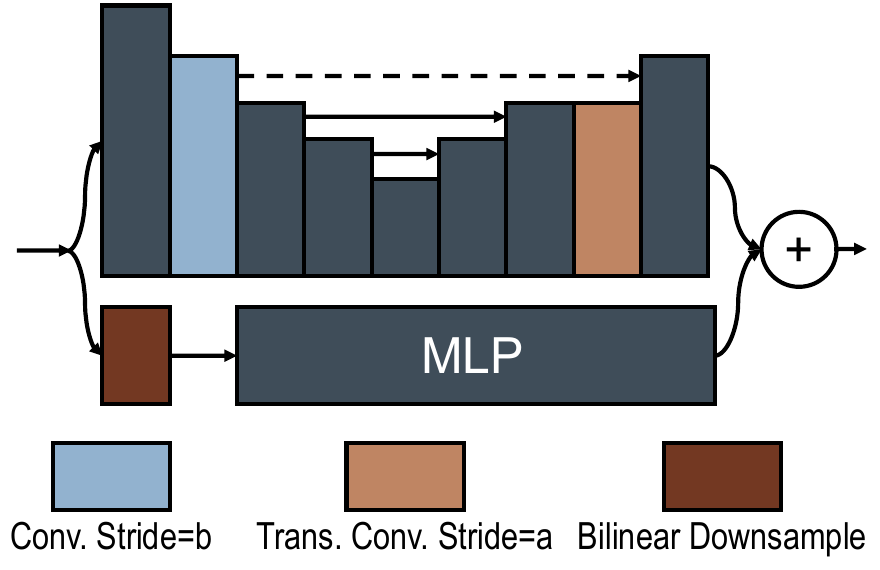}
        \caption{Preprocessor structure.}
        \label{fig:preproc}
    \end{subfigure}
    \caption{Overall encoding and decoding process with the proposed scheme: A standard codec is wrapped by a  neural wrapper with switchable weights.}
    \vspace{-4mm}
\end{figure*}

We design the coding framework shown in Fig.~\ref{fig:arch}. It involves an optional neural preprocesser (Fig.~\ref{fig:preproc}), a standard video codec, and an efficient neural postprocessor (Fig.~\ref{fig:post}).
Instead of using a fixed pair of preprocessor and postprocessor, we propose to employ multiple pairs, pretrained to resample the original video frame (with resolution $H \times W$) at different scales $r = \frac{a}{b}$, producing a \textit{bottleneck} video with resolution $\tilde H\times \tilde{W}=aH/b \times aW/b$. 
The encoder signals two video-level flags to the decoder, including the option of preprocessor (neural, linear, or none), and the downsampling mode  $k \in \{1, 2, 3, 4\}$ indicating the best $r_k$ chosen from rate-distortion optimization (RDO). The decoder loads the pre-trained pre-shared postprocessor model parameters and selects corresponding resampler based on the signaled mode.

 The neural preprocessor module is based on a standard UNet structure~\cite{ronneberger2015u} and a parallel 3-layer multilayer perceptron (MLP) ($1 \times 1$ convolution in other words), based on the design principle in \cite{guleryuz2021sandwiched}. The UNet supports different downsampling scales by configuring the highlighted layers in Fig.~\ref{fig:preproc} with different convolution strides. 
For the example shown in Fig.~\ref{fig:preproc}, to achieve $r = \frac{a}{b} = \frac{2}{3}$, the light blue layer uses $stride = 3$ for 3x downsampling, and the orange layer uses $stride = 2$ for 2x upsampling. We use $r \in \{\frac{1}{1}, \frac{1}{2}, \frac{2}{3}, \frac{1}{4}\}$ in our experiments. Note that the preprocessing is optional, and can be replaced by downsampling with a linear filter. We detail the postprocessing in Sec.~\ref{sec:post}.


\begin{figure}[t]
    \centering
    \includegraphics[width=0.7\linewidth]{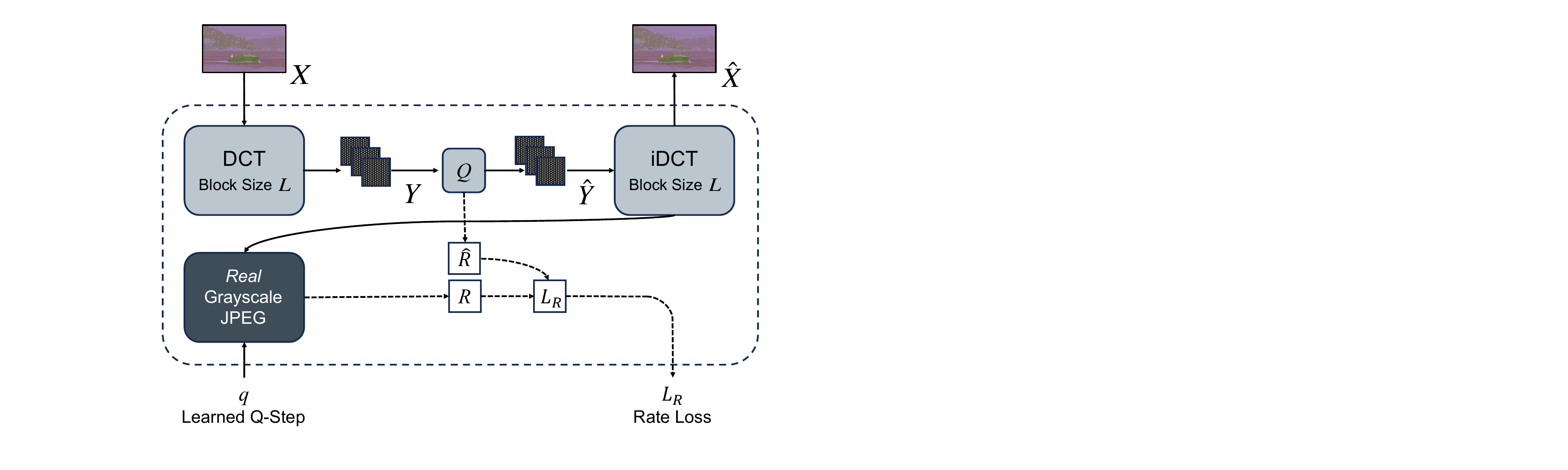}
    \caption{Extended JPEG Proxy for end-to-end training: The input $X$ is the 3-channel frame produced by the preprocessor.}
    \label{fig:proxy}
\vspace{-3mm}
\end{figure}

\vspace{-2mm}
\subsection{Efficient Post-Processor}
\label{sec:post}
\begin{figure*}
    \centering
    \includegraphics[width=0.9\linewidth]{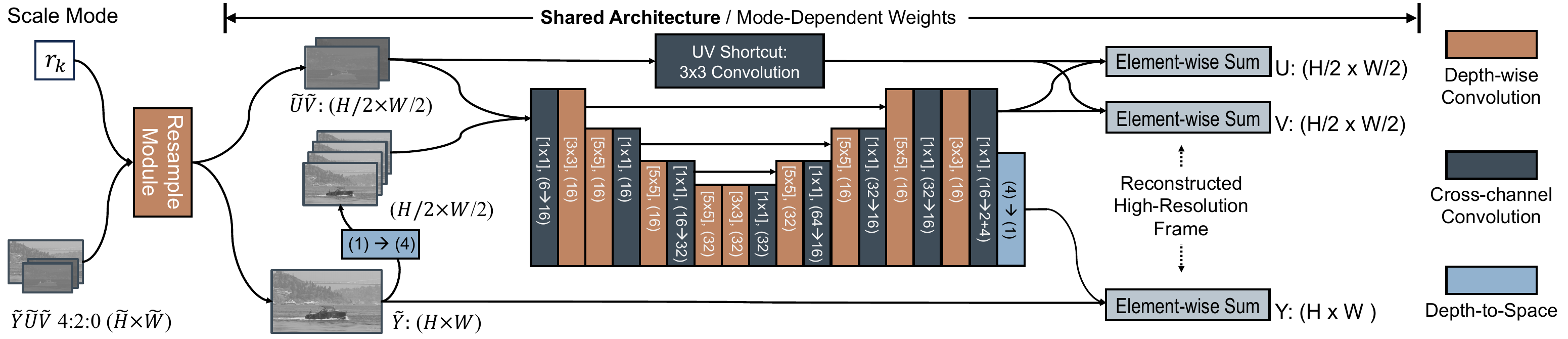}
    \caption{Structure of efficient post-processor. Layers are labeled with format [kernel size], (input / output channels) or (input $\rightarrow$ output channel) if there are different channel numbers for the input and output.}
    \label{fig:post}
    \vspace{-2mm}
\end{figure*}

The key component of the proposed video coding scheme is an efficient neural postprocessor that upsamples videos according to different upsampling ratio $r_k$, shown in Fig.~\ref{fig:post}. To universally support different bottleneck video resolutions, the input 4:2:0 frames represented by $\tilde{Y},\tilde{U}, \tilde{V}$ at resolution $\tilde{H} \times \tilde{W}= r_k H \times r_k W$
first go through a resampling module which resample them to $H\times W$ using a bilinear filter. We then apply a pixel shuffling \textit{space-to-depth} operation to the luma channel $\tilde{Y}$, organizing it into 4 channels of size $H/2 \times W/2$. The resulting 4 channels are next concatenated with $\tilde{U}$ and $\tilde{V}$, yielding a 6 channel signal that is sent to the neural postprocessor. The neural postprocesors for different downsampling ratios share the same architecture, with model parameters changed based on the signaled mode. This approach helps to simplify hardware implementation of the neural network. We practice the following principles for efficient neural network designs.

First, we recognize that cross-channel large-kernel convolution is costly in computation. Inspired by~\cite{howard2017mobilenets}, for any convolution layer that has a kernel size bigger than $1\times 1$, we only do per-channel convolution (denoted as depth-wise convolution). We then use a cross-channel $1\times 1$ convolution to combine different channels. This design achieves a good trade-off between the model capacity of the neural network and the model parameter count and complexity.

Second, we use a UNet-like~\cite{ronneberger2015u} structure to enable multi-resolution processing. Because the spatial dimensionality in intermediate layers is reduced, this structure also reduces the average MAC per pixel.

Third, we observe that learning to recover the residuals is easier.  To address this issue, we design a shortcut structure to bypass the input signal to the output. The postprocessor UNet is then learning just a residual signal. The shortcut operator functions differently on the three bottleneck image channels. As shown in Fig.~\ref{fig:post}, the UNet takes all six channels as input and generates six-channel residuals. The four channels associated with $\tilde{Y}$ are then converted back to a single channel through a depth-to-space module, which  is then added to the input $\tilde{Y}$ to produce the reconstruction of the luma channel $Y$. $\tilde{U}$ and $\tilde{V}$ go through a learned $3\times 3$ cross-channel convolution layer, which are then added to their corresponding residuals from the UNet to produce the chroma $U$ and $V$. With this design, the preprocessor is encouraged to make $\tilde{Y}$ similar to the low-resolution $Y$ channel. This also helps the wrapped standard video codec conduct more efficient motion compensation on the $\tilde Y$ channel. We have also experimented using actual luma signal for $\tilde Y$ (i.e. directly downsample $Y$ to generate $\tilde Y$, but use a neural preprocessor on $U, V$ to generate ${\tilde U}, {\tilde V}$), but found that training a pre-processor to generate $\tilde Y$ gives slightly better performance.

With the proposed design, we are able to reduce the complexity to 516 MACs~/~pixel (measured using thop~\cite{thop}), and number of parameters to 8.2 K. It makes our system immediately adoptable on existing desktop platforms. It can also be potentially accelerated by AI accelerators on mobile devices.

\vspace{-2mm}
\subsection{Jointly Optimize Neural Wrapper with Codec Proxy}

In the neural wrapper, the bottleneck video should be rate-constrained in the context of a common transform coding scheme. After coding with a standard codec, the bottleneck videos are likely to carry quantization noise and other coding artifacts. Hence the post-processor should also be trained with simulated artifacts.

Following the design in \cite{guleryuz2021sandwiched}, we construct a differentiable JPEG proxy to apply constraints, estimate coding bit-rate, and simulate coding artifacts in the end-to-end training of the neural wrapper. Unlike \cite{guleryuz2021sandwiched}, we randomly set the block size for DCT  to simulate variable block size coding in a video codec.  Given a bottleneck frame $X$ of three channels, we first clip its values to the range $[0, 2^d - 1]$, where $d \in \{8, 10\}$ is the target bit-depth. We then partition $X$ into $L \times L$ blocks, and apply $L \times L$ DCT over each block to create coefficients $Y$. During training, $L$ is randomly selected from $\{4, 8, 16, 32\}$. We apply quantization on $Y$ and produce $\hat{Y}$ using a learnt quantization stepsize $q$.  We then perform inverse DCT on $\hat{Y}$ to generate the reconstructed frame $\hat{X}$. During gradient backpropagation, we simply bypass the quantizer.

We estimate the bit-rate by a scaled sum of logarithm of the absolute values of the coefficients, \textit{i.e.}
\begin{equation}
\label{eq:lr}
\begin{split}
    &L_R = a \sum_{c,i,j} \log (1+|Y_{c,i,j}|/q),\\
    &a = \text{SG} \left( \frac{R} {\sum_{c,i,j} \log (1+|Y_{c,i,j}|/q)}\right),
\end{split}
\end{equation}
where $i \in [0, H-1], j \in [0, W-1]$ and $c \in {0, 1, 2}$ are indices for the spatial pixel positions and channels of the bottleneck coefficients $Y$. 
The scaling factor is calculated such that $L_R$ equals the  bit-rate obtained by a JPEG coder on the reconstructed image $\hat X$. Specifically, given $\hat{X}$, we send each of its channel to a real grayscale JPEG\footnote{JPEG is used for convenience and speed during training. Our approach is easily generalized to use other codecs such as HEVC.}, and use the corresponding quality factor (qf) calculated from $q$ to encode. We use $\text{qf} = 100 - \frac{1}{32}(100q- 50)$, the qf when the DC quantization stepsize is $q$.
The real JPEG gives out a bit-rate $R$, and we use it to calculate $a$. Note that to have gradients back-propagated to $Y$ and $q$, we need to use a stop-gradient function $\text{SG}()$ onto the scaling factor $a$ when we implement the loss. Meanwhile, since JPEG only supports 8-bit images, when our target bit-depth $d = 10$, we take the most significant 8 bits of the value at each pixel when we run the real JPEG.

The quantization on the coefficients simulate the common lossy coding artifacts. Hence, the post-processor learns to be resilient to such degradations. With $L_R$, the pre-processor also learns to generate valid bottleneck  images with a rate constraint.

\subsection{Training}

For the neural wrappers, we end-to-end train each pair of pre- and post-processor for a given $r$ with the codec Proxy, using the rate distortion loss function with a parameter $\lambda$ as,
\begin{equation}
    L = L_D + \lambda L_R = D(x, \hat{x}) + \lambda L_R, 
\end{equation}
where we use Mean Squared Error (MSE)  in the YUV space as the distortion metric $D$, and we use the $L_R$ as defined in Eq.~(\ref{eq:lr}). $x$ and $\hat{x}$ denote the original frames and reconstructed frames in YUV, respectively. For the post-processor only mode, we replace the preprocessor with a bilinear downsampling filter and train the post-processor only with the same loss function.

We train the model using images in the CLIC dataset~\cite{CLIC2020}. We use random crops of $256\times 256$ in training. We convert the images in RGB color space to YUV 4:2:0. Our model supports different source bit-depths by normalizing the input always to $[-1, 1]$. It also supports different  bit-depth for the bottleneck, by clipping the output to range $[-1, 1]$, shifting and rescaling it to $[0, 1]$, and multiplying it by the designated bit-depth  (255 for 8-bit and 1023 for 10-bit). Through empirical study we found that using $\lambda=16$ and bit-depth $d=10$ for the bottleneck channels gives the best result for, all bit-rate ranges (both 8-bit and 10-bit sources) and all wrapped video codecs.

We train a total of 4 pairs of neural processor and postprocessor for 4 different downsampling factors. We also train 4 additional neural postprocessors corresponding to bilinear downsamplers. All models are trained from scratch.
\section{Experiments}
\subsection{Settings}
Our method is evaluated on the UVG dataset (4K, YUV 4:2:0, 8-bit, first 120 frames)~\cite{mercat2020uvg} and AOM CTC (Class A1, 4K, YUV 4:2:0, 10-bit, all 130 frames)~\cite{AOMCTC}.
One advantage of our method is that once the model pair for a particular downsample factor $r$ is trained, it can be applied to various standard codecs (\textit{e.g.} HEVC, VVC). We use the following settings:

\begin{itemize}[leftmargin=*]
\item \textbf{HEVC}. HEVC encodes the original 4K 4:2:0 videos directly. We use x265 (8-bit for the UVG dataset and 10-bit for AOM CTC) at the medium preset with the default configuration~\cite{x265presets}.



\item  \textbf{VVC}. We use VVenC~\cite{VVenC} and VVdeC~\cite{VVdeC} to encode the 4K videos directly. We use YUV 4:2:0 10-bit configurations with the medium preset.


\item \textbf{Ours (HEVC) / Ours (VVC)}. With the framework in Fig.~\ref{fig:arch}, the encoder processes a sequence multiple times with the following options: 1) enabling neural wrappers with $r \in \{\frac{1}{1}, \frac{1}{2}, \frac{2}{3}, \frac{1}{4}\}$; 2) only using post-processor with $r \in \{\frac{1}{1}, \frac{1}{2}, \frac{2}{3}, \frac{1}{4}\}$, and 3) coding the original frame directly with the codec (HEVC or VVC). The bottleneck videos are always coded using 10-bit YUV 4:2:0 format. The encoder selects the option that gives the highest PSNR for each bit-rate point and for each sequence. We use 4 bits for the whole video to signal the modes.  When calculating the BD-rate, we use HEVC (VVC) as the anchor when we our scheme wraps the (VVC), respectively.
\end{itemize}

\begin{figure}[t]
    \centering
    
    \begin{subfigure}{0.8\linewidth}
        \centering
        \includegraphics[width=0.9\linewidth]{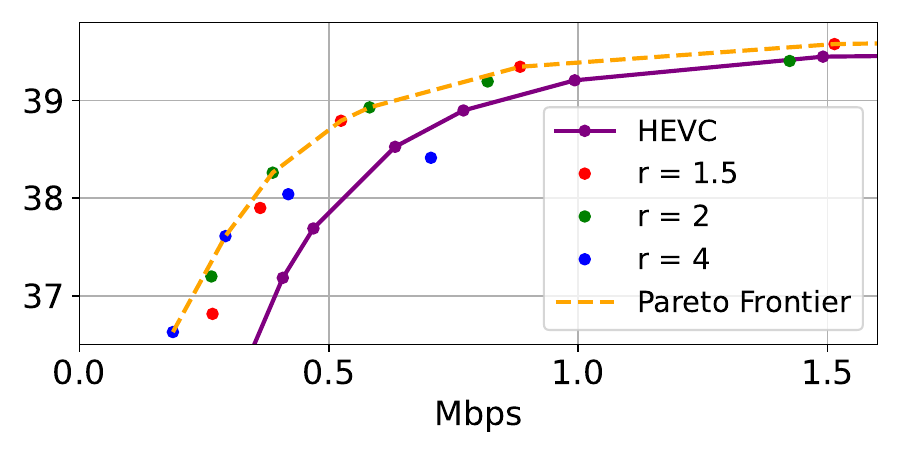}
        \vspace{-2mm}
        \caption{Honeybee (HEVC)}
    \end{subfigure}
    
    \begin{subfigure}{0.8\linewidth}
        \centering
        \includegraphics[width=0.9\linewidth]{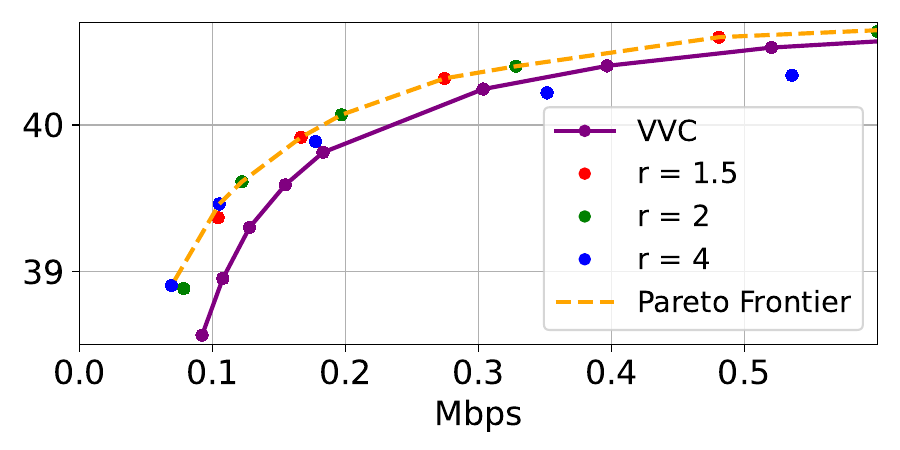}
        \vspace{-2mm}
        \caption{FlowerFocus (VVC)}
    \end{subfigure}
    
    \caption{Illustration of Pareto Frontier by combining  R-D curves resulting from all downsampling ratios on two typical sequences in the UVG dataset (only postprocessor is used).}
    \label{fig:pareto}
    \vspace{-4mm}
\end{figure}

We average YUV PSNR by the weight $6:1:1$. We obtain the R-D curve for each sequence by calculating a Pareto frontier from all achievable R-D points using all available modes. An example of the R-D Pareto frontier is shown in Fig.~\ref{fig:pareto}, using the post-processor-only setting with HEVC and VVC, respectively. As shown, different $r$ has different advantage bit-rate regions, and lower $r$ is necessary to achieve improvements at higher bit-rates.

\subsection{Effectiveness of Neural Wrappers}

\begin{table}[t]
    \centering
    \footnotesize
    \caption{Average BD-Rate on the UVG dataset with different settings. We use HEVC as the anchor.}
    \centering
        \begin{tabular}{lc}
        \toprule
        Method & BD-Rate (HEVC)  \\
        \midrule
        Bilinear & -7.4 \%  \\
        Lanczos-4 & -11.0 \%  \\
        Only Post Proc. & -18.1 \%  \\
        Wrapper & -19.9 \%   \\
        Only Post + Wrapper & -22.6 \% \\ 
        \bottomrule
        \end{tabular}
    \label{tab:ablation}
\end{table}


\begin{figure}[t]
    \centering
    \begin{subfigure}{0.8\linewidth}
        \centering
        \includegraphics[width=0.9\linewidth]{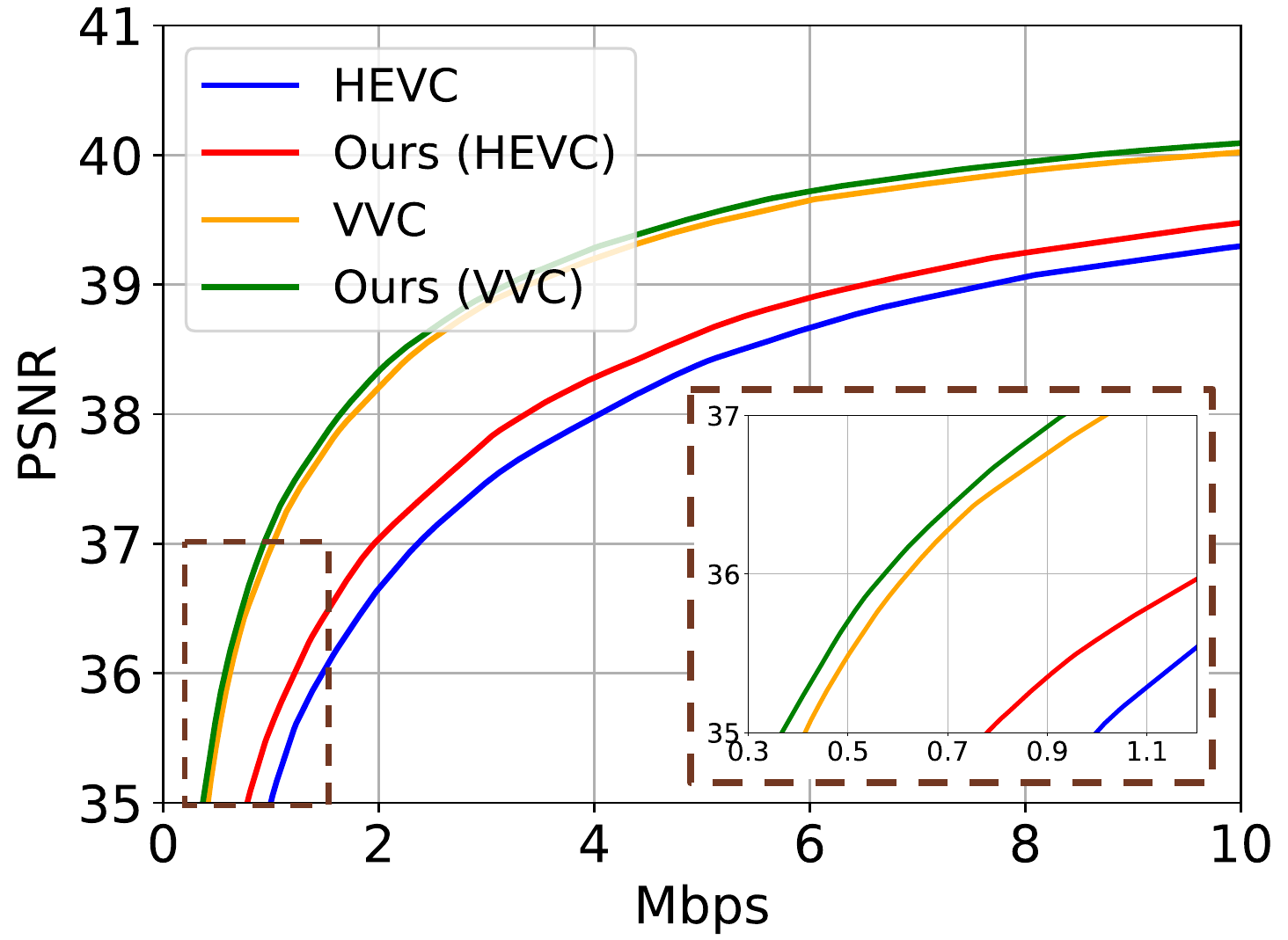}
    \end{subfigure}
    \caption{Rate-distortion curves averaged over all videos in the UVG dataset. We achieve -22.6\% BD-Rate over HEVC and -9.3\% over VVC.
    }
    \label{fig:rd}
    \vspace{-4mm}
\end{figure}

We compare the proposed scheme with the following baselines to study the effectiveness of the neural pre- and post-processing,

\noindent 1) \textbf{Linear Filters}. We use \textit{Bilinear} or \textit{Lanczos-4} filter to first downsample the video, code using HEVC, and upsample using the same filter. The encoder choose from 4 options.

\noindent 2) \textbf{Only Post Proc.}. We use bilinear downsampling and replace the upsampling filter with the neural postprocessor. The encoder choose from 5 options, including 4 downsampling ratios and HEVC coding the original directly.

\noindent 3) \textbf{Wrapper}. We use the end-to-end trained pre- and post-processors pairs. The encoder choose from 5 options.

\noindent 4) \textbf{Only Post + Wrapper}. We combine the above two with RDO. The encoder chooses from 9 options.

\noindent As shown in Table~\ref{tab:ablation}, the proposed neural pre- and post-processors can further boost the performance by 15.2\% over the baseline using the linear filters with HEVC. 

We plot the average R-D curves on the UVG dataset in Fig.~\ref{fig:rd}. With RDO from different modes, on the UVG dataset, we achieve -22.6\% BD-Rate over HEVC, and -9.3\% over VVC. We also achieve -6.4\% BD-Rate over VVC on AOM CTC Class A1 with the proposed scheme.



\begin{table}[t]
    \centering
    \footnotesize
    \caption{Performance and complexity comparison with the sandwiched codec~\cite{isik2023sandwiched}.}
    \begin{tabular}{c|c|c}
    \toprule
        Method  & BD-Rate & kMACs / pixel  \\
    \midrule
        Ours    & -18.9\%  & 0.5 \\
        Sandwiched Codec~\cite{isik2023sandwiched} & -19.5\%  & 471 \\
    \bottomrule
    \end{tabular}
    \label{tab:compare_sandwich}
    \vspace{-4mm}
\end{table}

In our work, we specifically target real-world applications on existing hardware platforms. To this end, we made non-trivial efforts to reduce the decoding complexity to 516 MACs/pixel, enabling real-time decoding of 4K videos. We compare the proposed method to the work~\cite{isik2023sandwiched} to illustrate the significance in complexity reduction. We implement~\cite{isik2023sandwiched} and train / test with the same configurations. For a fair comparison, we used both methods with the combination of the wrapper ($r=\frac{1}{2}$) and the original codec (HEVC), and benchmarked them on the UVG dataset. As shown in Table~\ref{tab:compare_sandwich}, with approximately 1000 times savings in complexity, our method achieves comparable R-D performance. We emphasize that the MACs / pixel directly links to hardware resources and power consumption. In comparison to end-to-end learned neural codecs which have complexity from 100K to 1M MACs / pixel, our work achieves more R-D gain with a fraction of the complexity the simplest codec has. We believe this is a significant advance in practical high-performance compression with neural tools.

\vspace{-4mm}
\subsection{Qualitative Analysis}
\begin{figure}[t]
    \centering
    \begin{subfigure}{0.49\linewidth}
        \centering
        \includegraphics[width=0.85\linewidth]{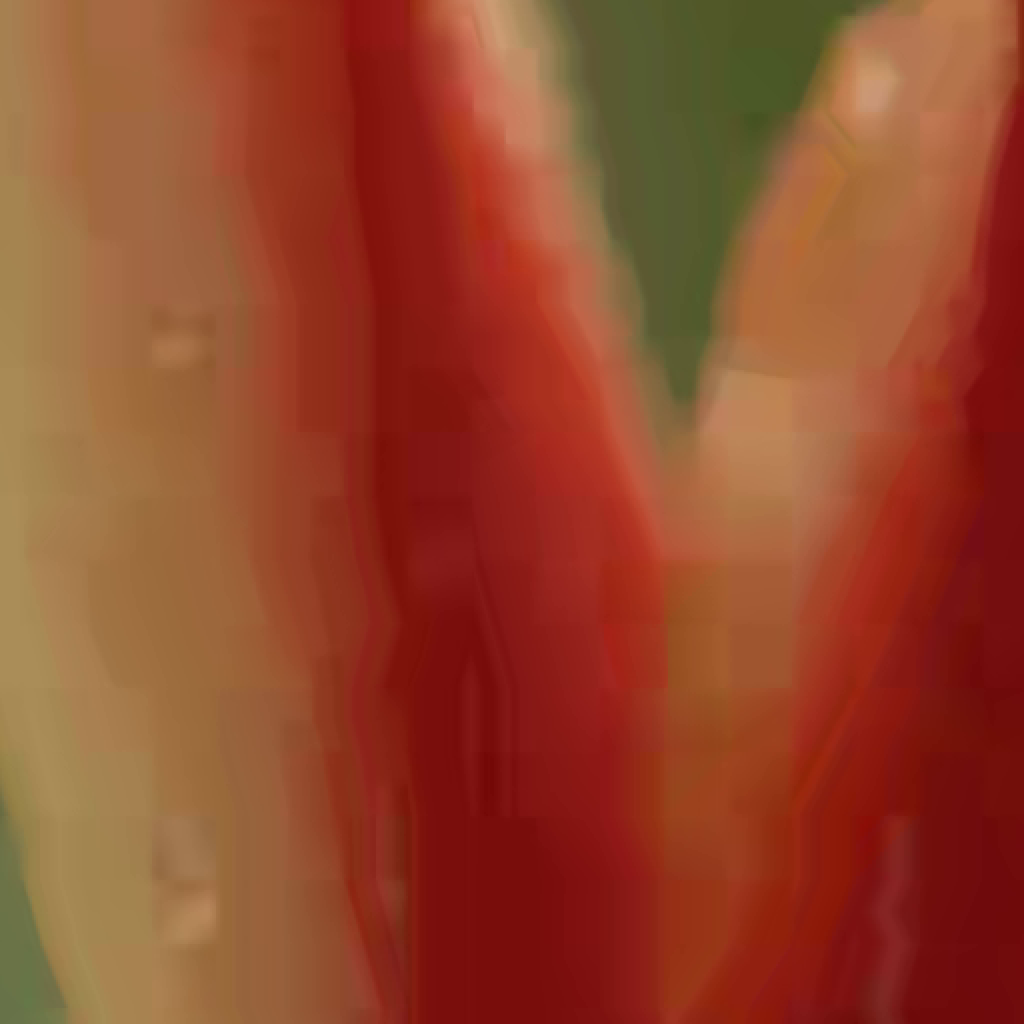}
        \caption{\footnotesize{HEVC, 3.0 Mbps}}
        \label{fig:flower_a}
    \end{subfigure}
    \begin{subfigure}{0.49\linewidth}
        \centering
        \includegraphics[width=0.85\linewidth]{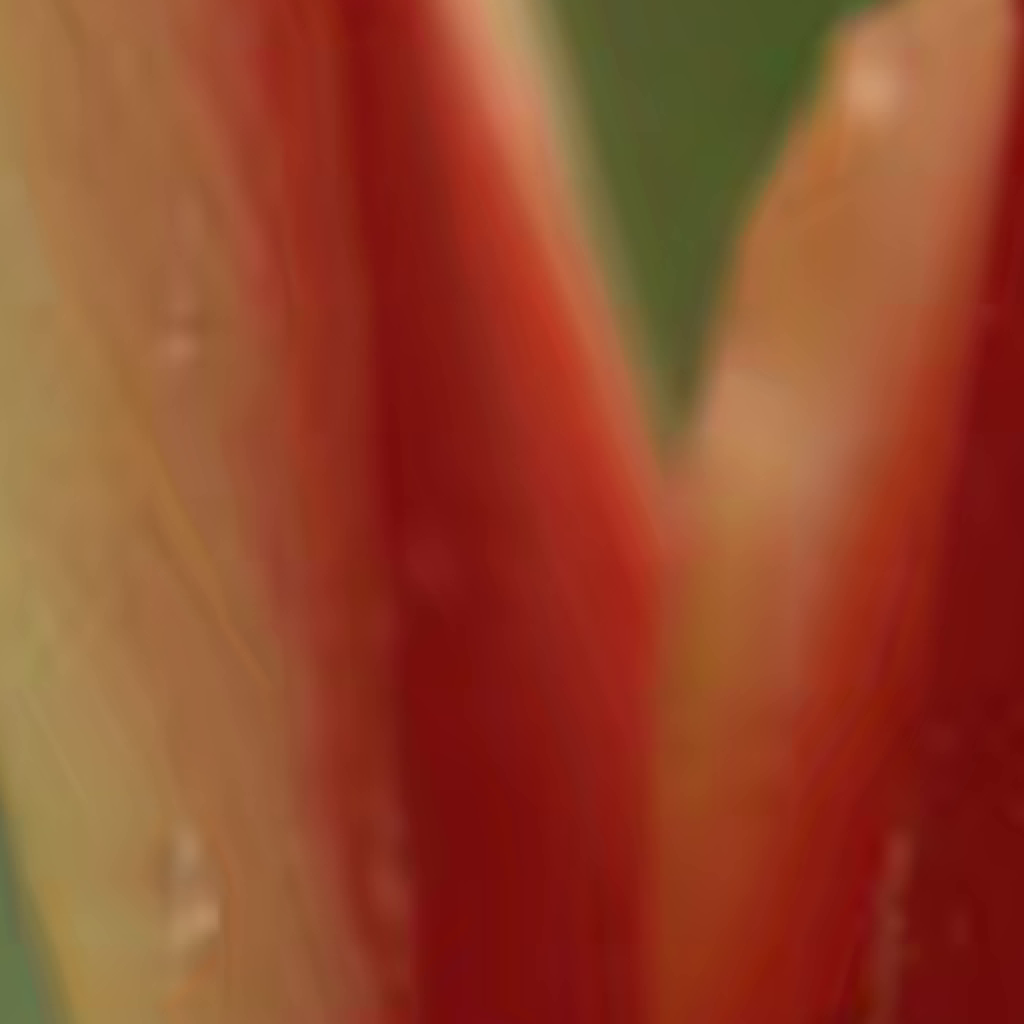}
        \centering
        \caption{\footnotesize{Ours, 2.9 Mbps}}
        \label{fig:flower_b}
    \end{subfigure}
    
    \begin{subfigure}{0.49\linewidth}
        \centering
        \includegraphics[width=0.85\linewidth]{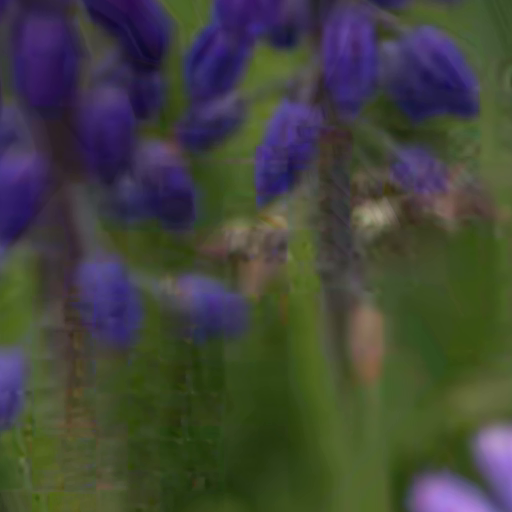}
        \caption{\footnotesize{HEVC, 4.3 Mbps}}
        \label{fig:honey_a}
    \end{subfigure}
    \begin{subfigure}{0.49\linewidth}
        \centering
        \includegraphics[width=0.85\linewidth]{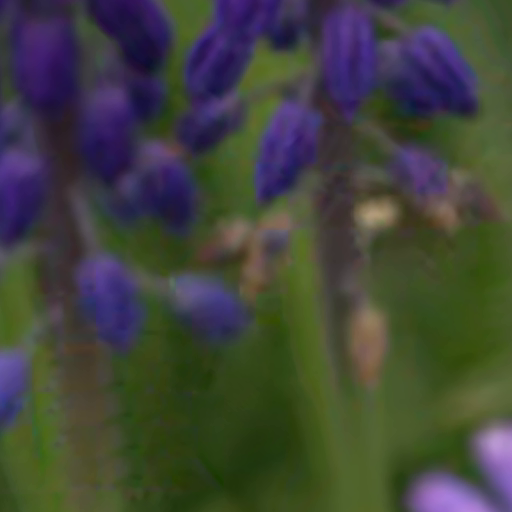}
        \caption{\footnotesize{Ours, 4.3 Mbps}}
        \label{fig:honey_b}
    \end{subfigure}
    \caption{Visual quality comparison on the \textit{FlowerFocus} and \textit{HoneyBee} sequences in the UVG dataset.}
    \label{fig:visual}
    \vspace{-4mm}
\end{figure}

Fig.~\ref{fig:visual} shows the visual comparison over HEVC with two testing sequences.
In Fig.~\ref{fig:flower_a} and~\ref{fig:flower_b}, due to quantization, HEVC causes annoying distortion on the edge of the petal and blurs the contour. Since our method encodes the video in lower resolution, effectively using a bigger base function with less aggressive quantization, the distortion is far less severe.
In Fig.~\ref{fig:honey_a} and~\ref{fig:honey_b}, because of error propagation in motion compensation, HEVC leaves out a trace of noisy area due to uncompensated residuals. We reduce this effect by making motion compensation easier at a lower resolution. Besides, our method also serves as a better postprocessing filter. Our method generally improves visual quality on this kind of natural scenes.

\vspace{-4mm}
\section{Conclusion}

In this paper, we propose a standard-compatible learned video coding framework that takes the advantage of neural networks for boosted R-D performance, while maintaining very low decoding complexity. With the proposed neural pre- and post-processors, our scheme achieves -9.3\% BD-Rate over VVC on the UVG dataset, and -6.4\% over VVC on AOM CTC Class A1. Our versatile framework  provides the flexibility of updating the pre-processor and post-processor  with better trained parameters or using  other  pre-processor and  post-processor architectures, to support a broader range of content types and application scenarios. For example, weights of the post-processor can be updated for specific video content, and pre-processors can be simplified for real-time encoding. 


\bibliographystyle{IEEEbib}
\bibliography{refs.bib}



\section*{Supplementary material}

# PART I
## Rate-Distortion Curves on UVG Dataset with HEVC

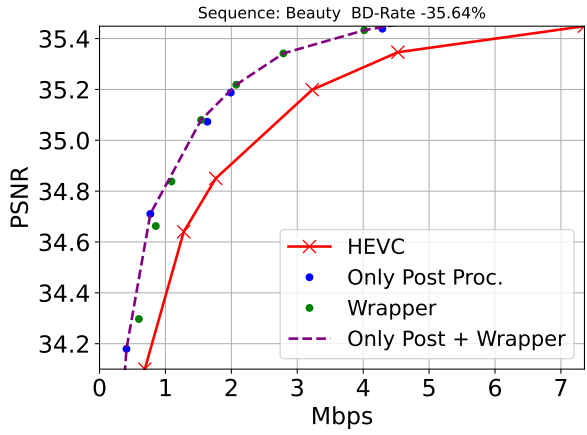

(a) Small range

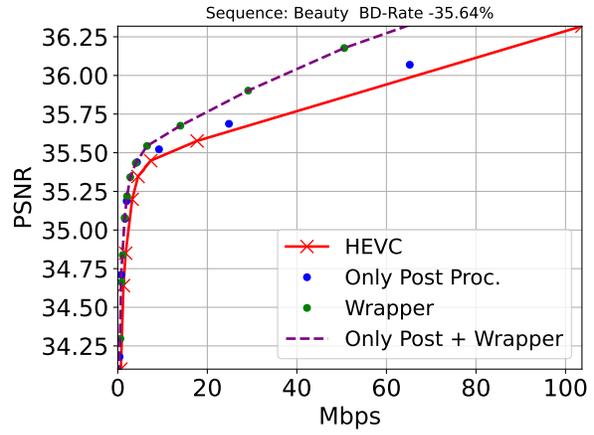

(b) Large range

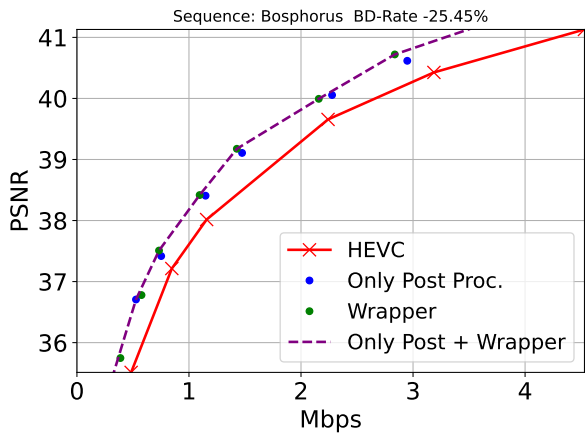

(a) Small range

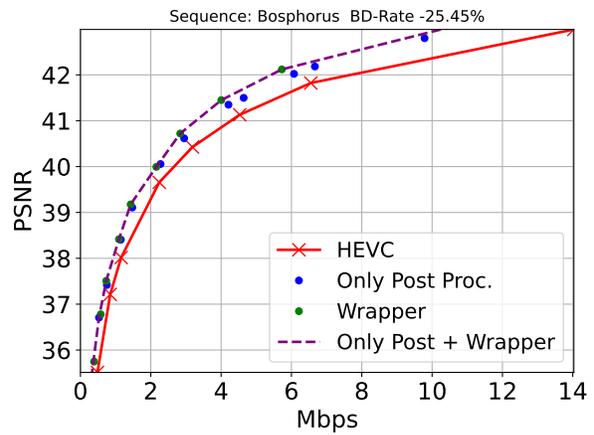

(b) Large range

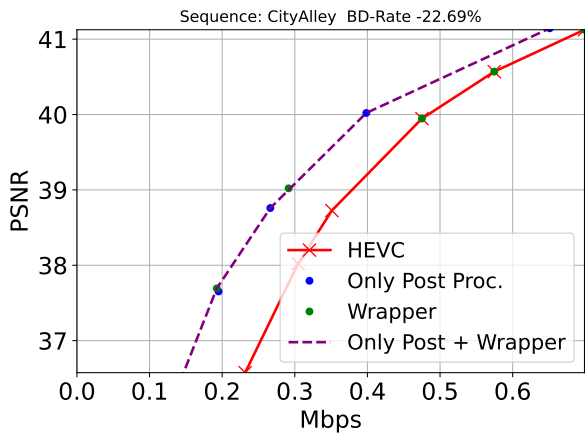

(a) Small range

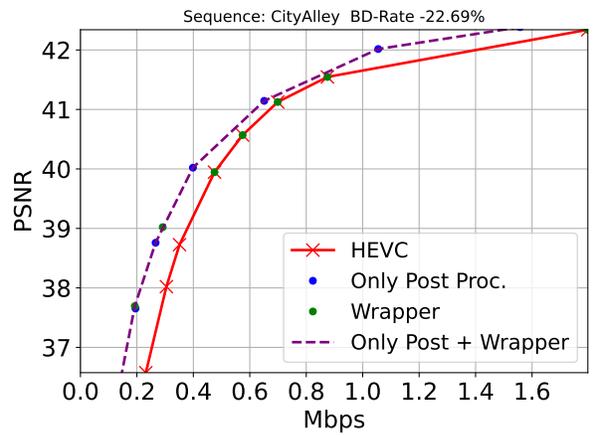

(b) Large range

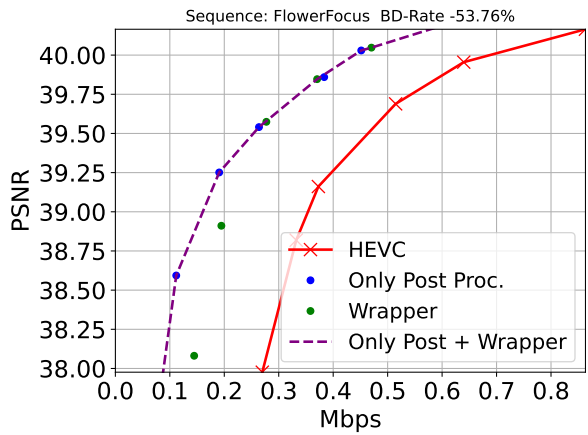
(a) Small range

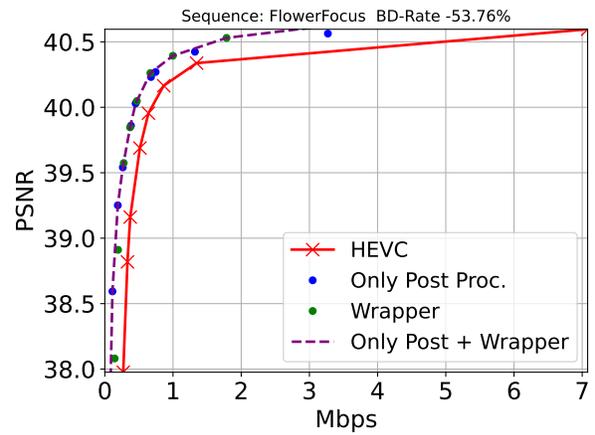
(b) Large range

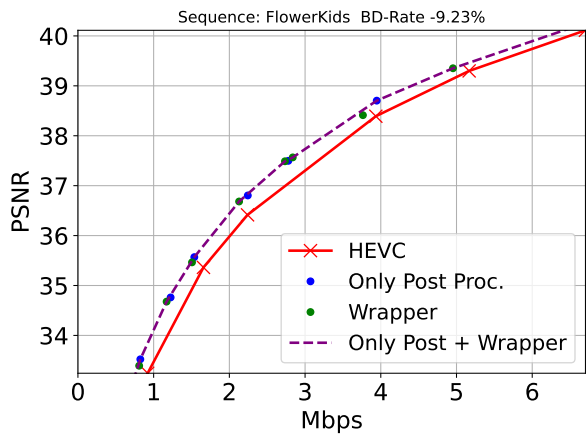
(a) Small range

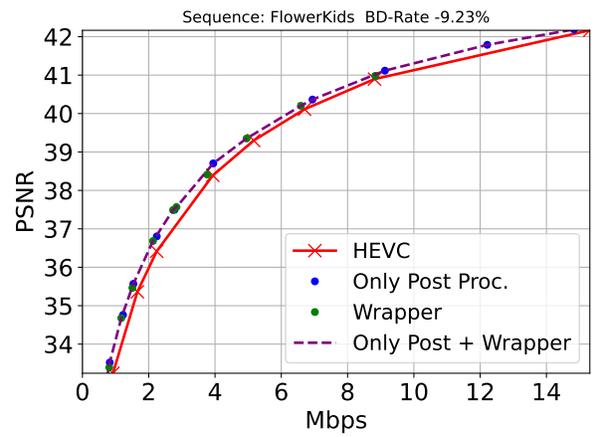
(b) Large range

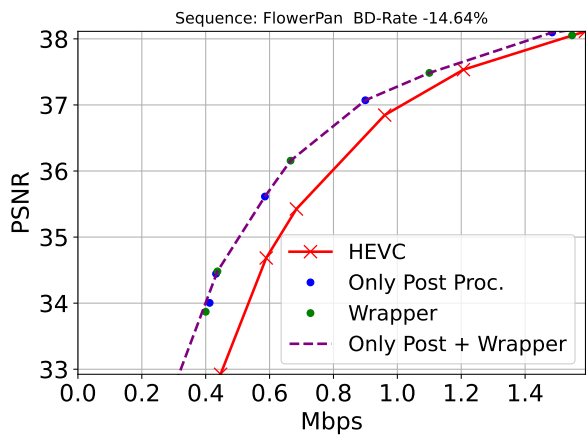
(a) Small range

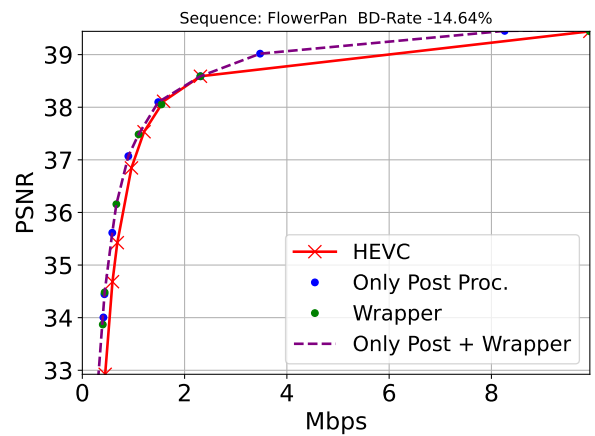
(b) Large range

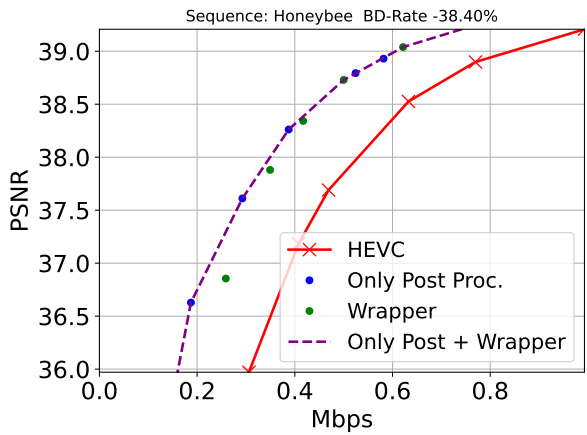
(a) Small range

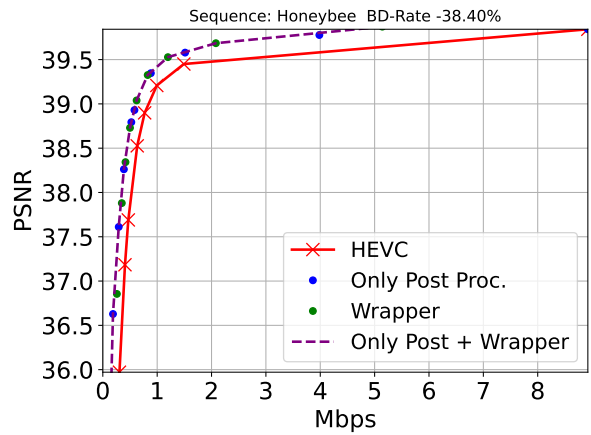
(b) Large range

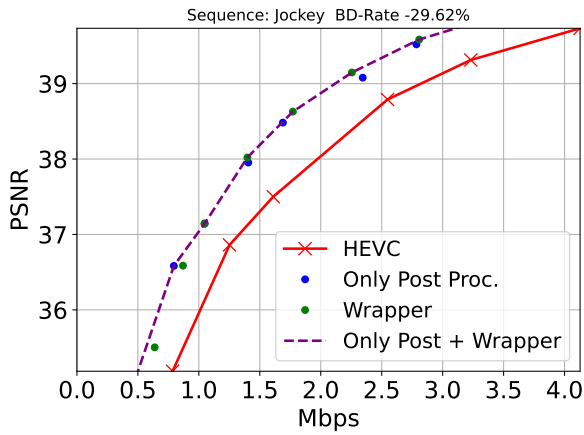
(a) Small range

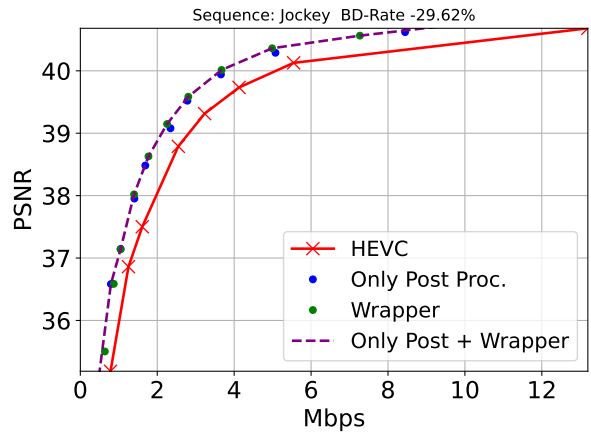
(b) Large range

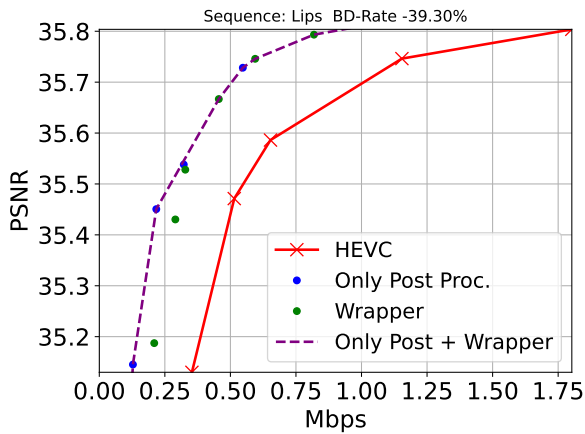
(a) Small range

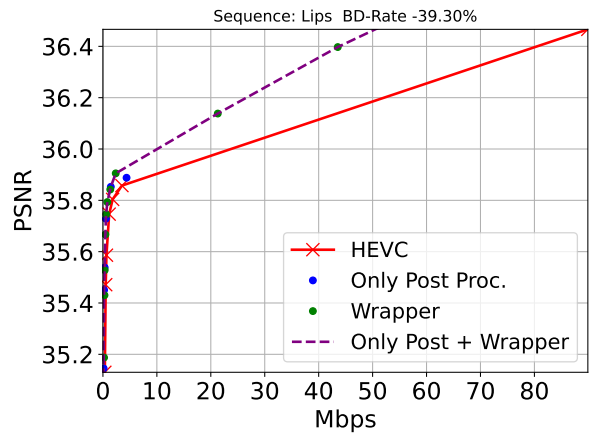
(b) Large range

4

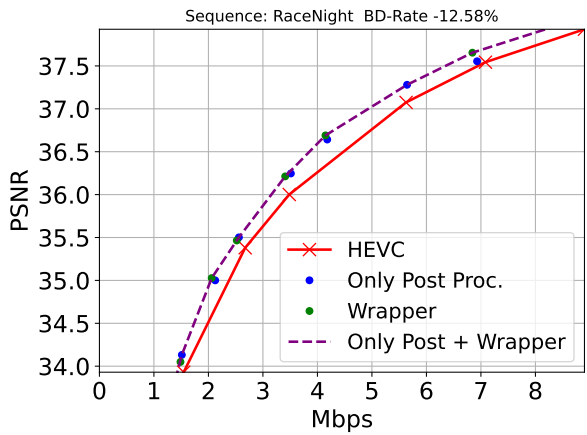

(a) Small range

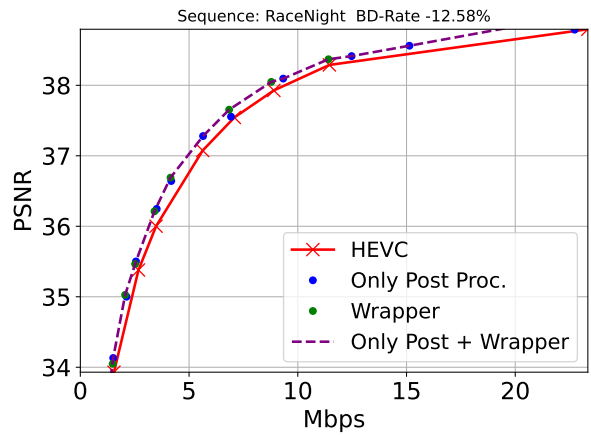

(b) Large range

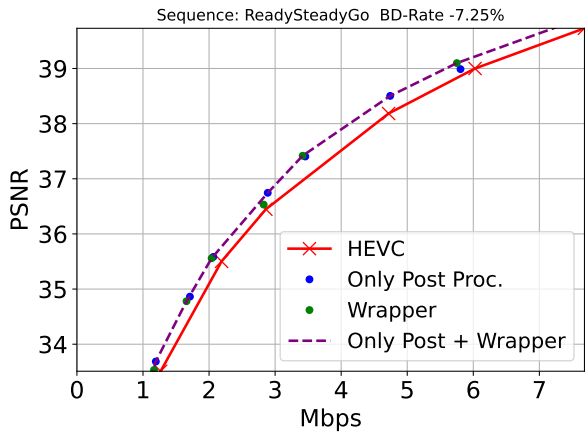

(a) Small range

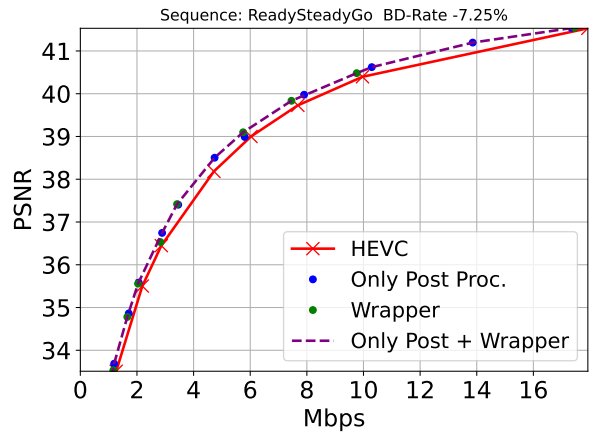

(b) Large range

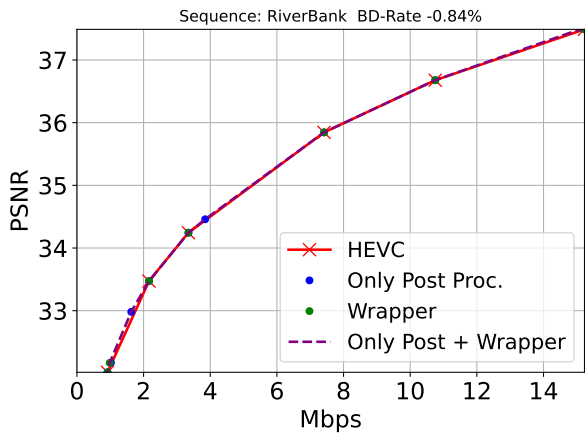

(a) Small range

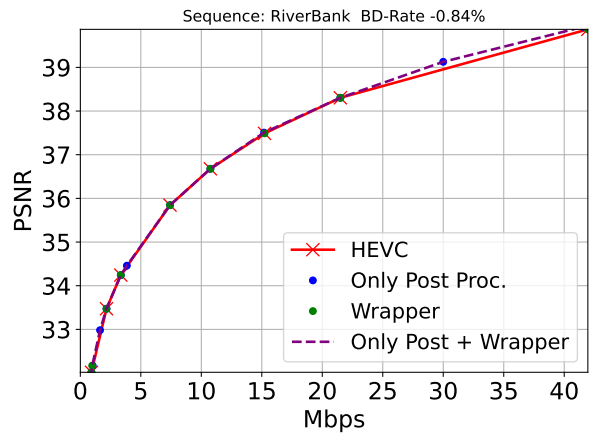

(b) Large range

5

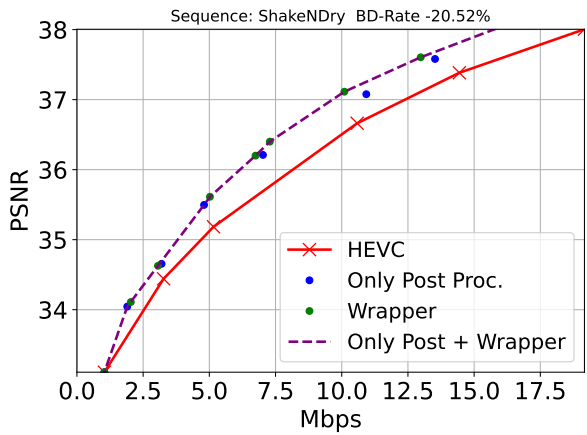
(a) Small range

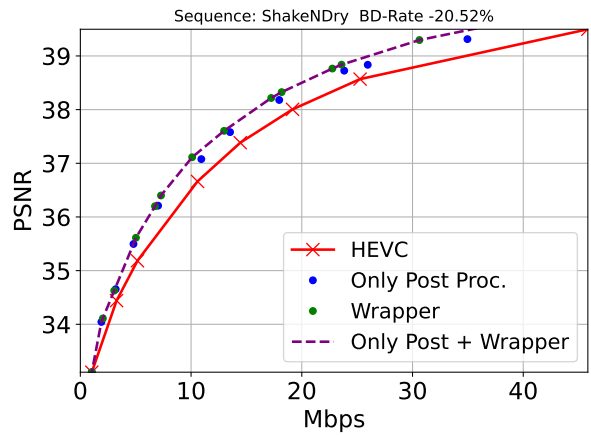
(b) Large range

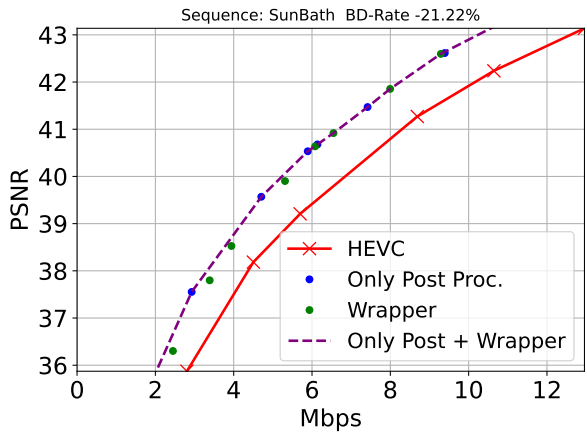
(a) Small range

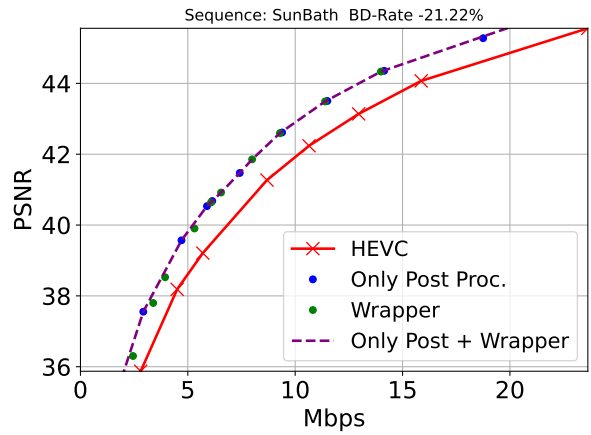
(b) Large range

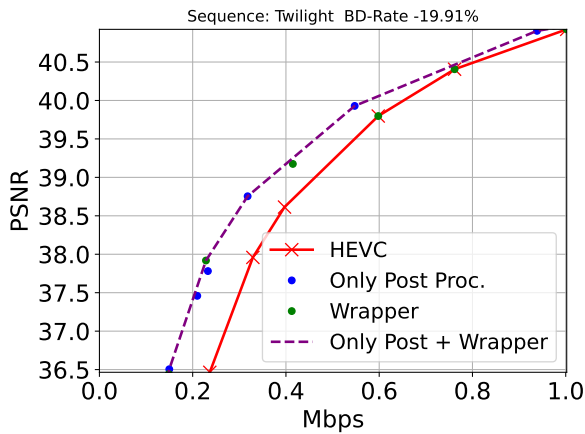
(a) Small range

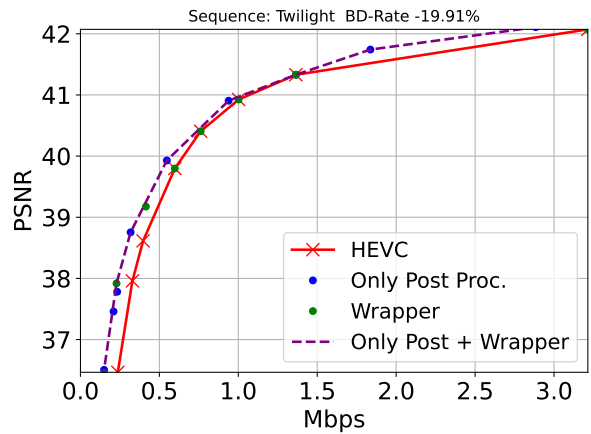
(b) Large range

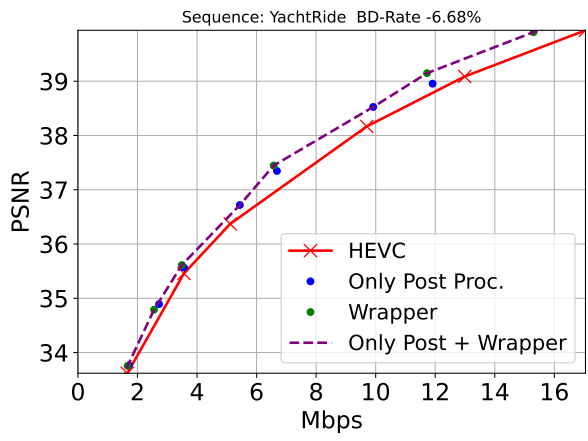
(a) Small range

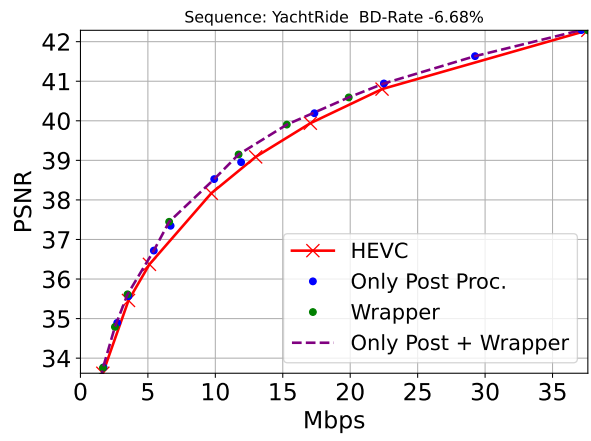
(b) Large range

# PART II
## Rate-Distortion Curves on UVG Dataset with VVC

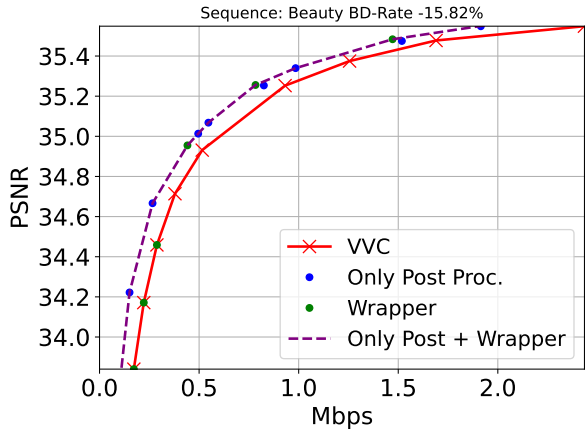

(a) Small range

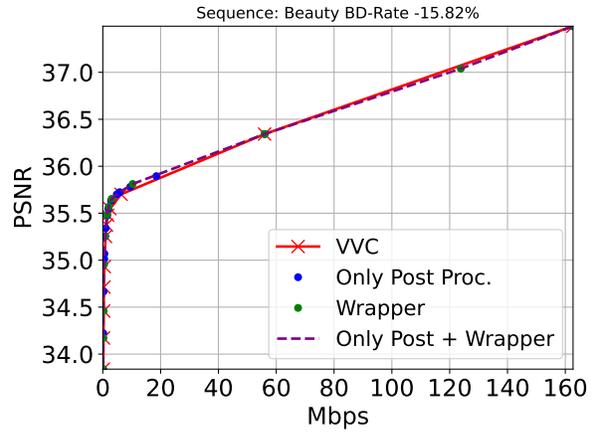

(b) Large range

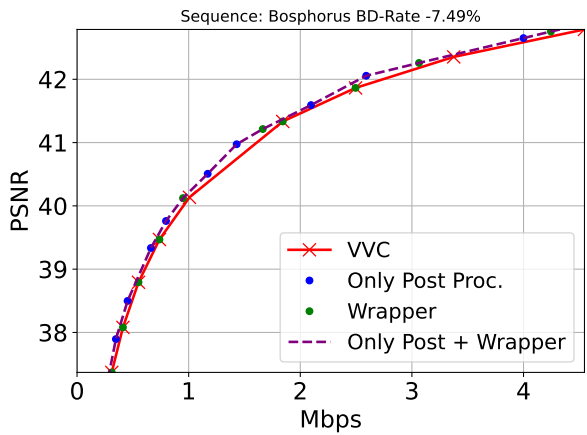

(a) Small range

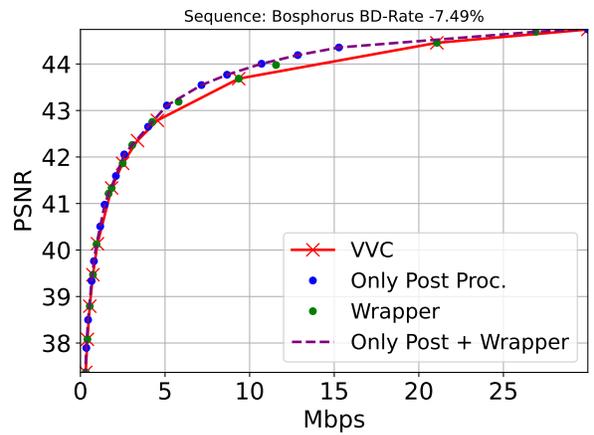

(b) Large range

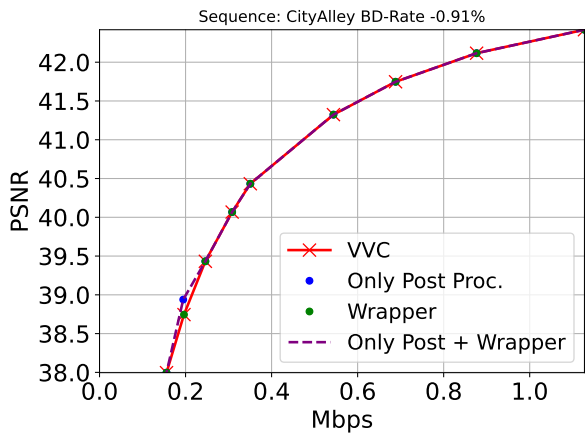

(a) Small range

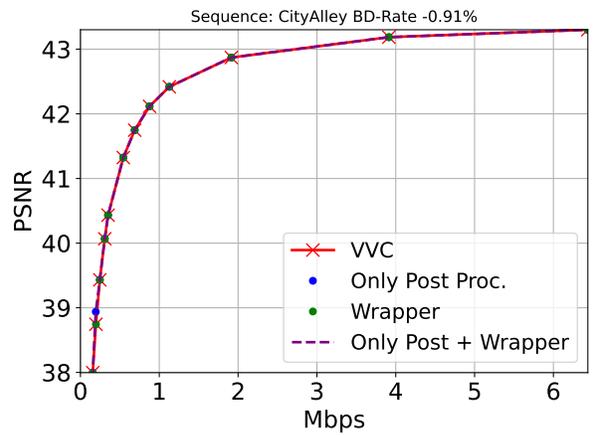

(b) Large range

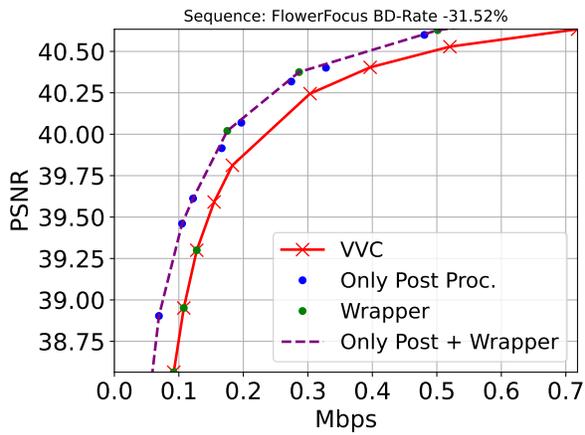
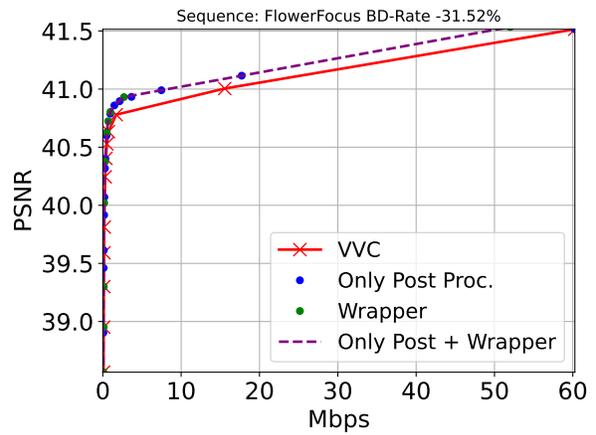

(a) Small range

(b) Large range

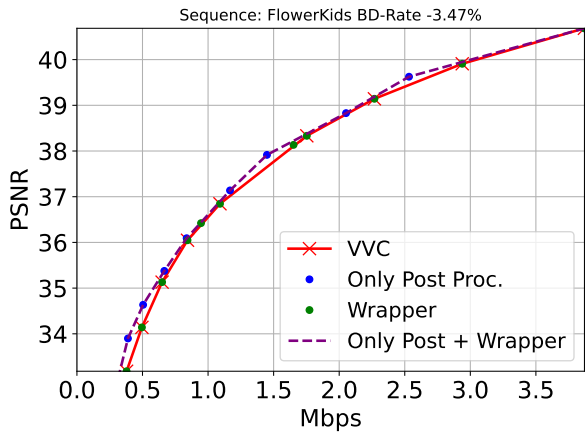
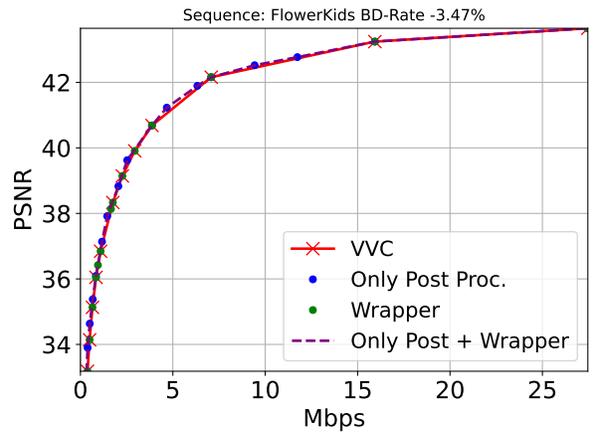

(a) Small range

(b) Large range

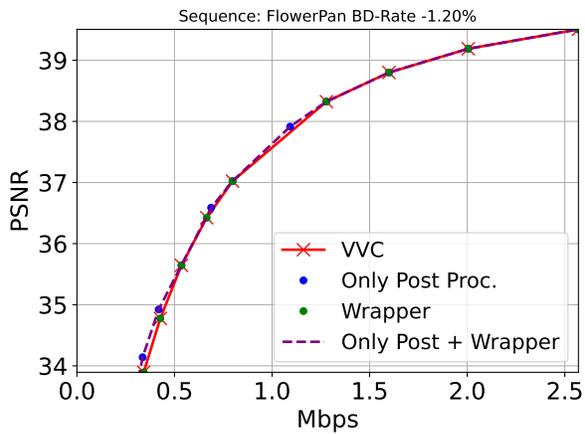
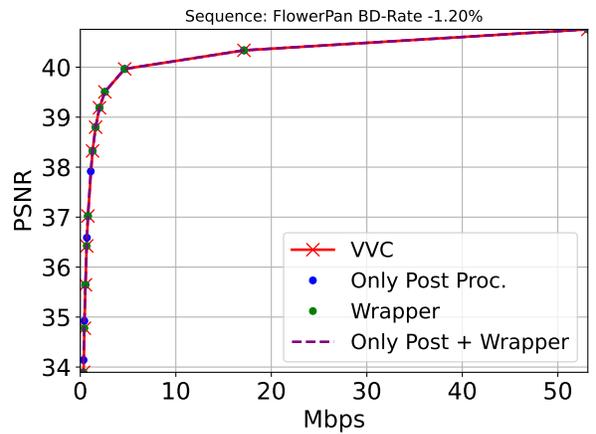

(a) Small range

(b) Large range

9

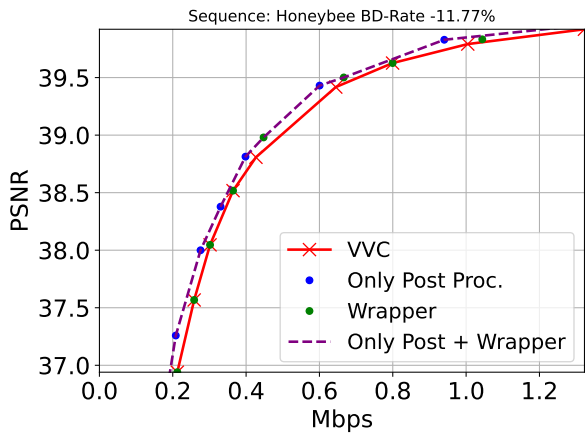
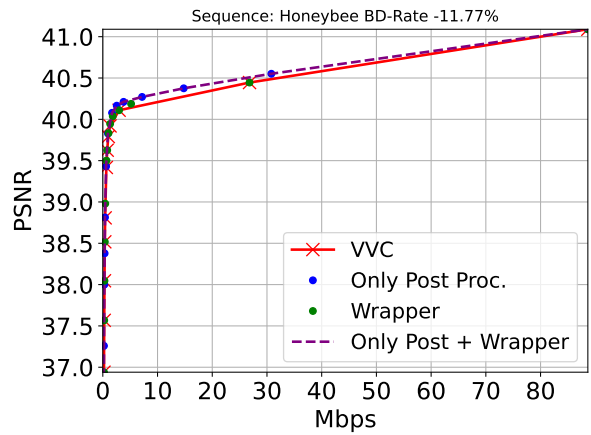

(a) Small range  (b) Large range

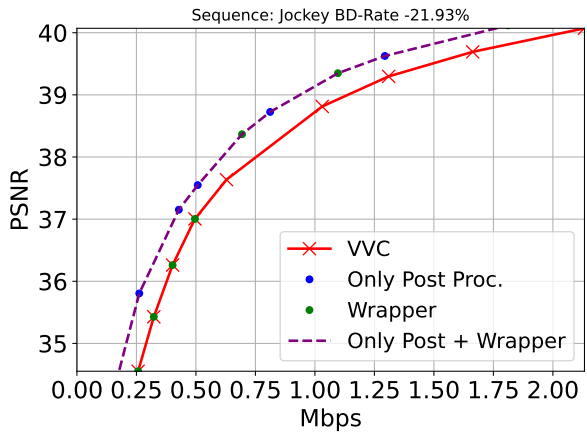
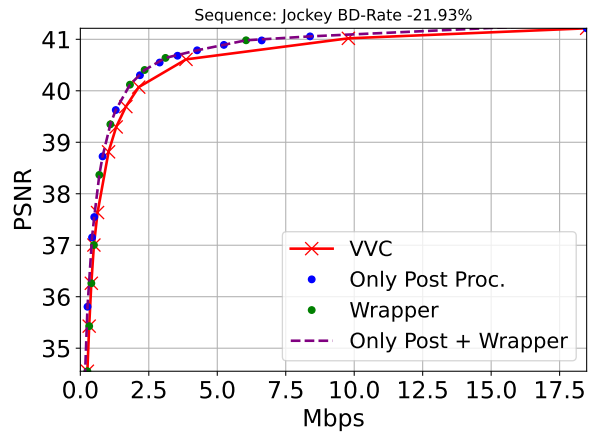

(a) Small range  (b) Large range

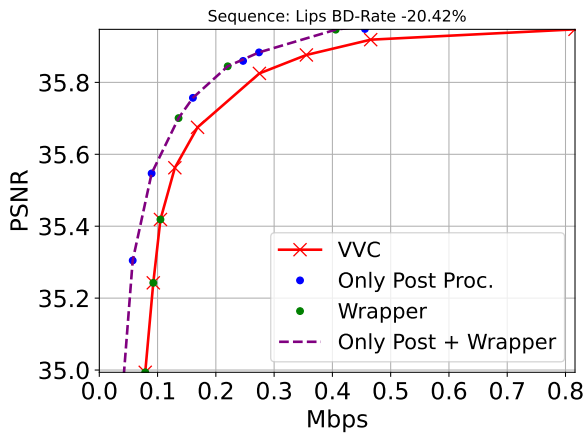
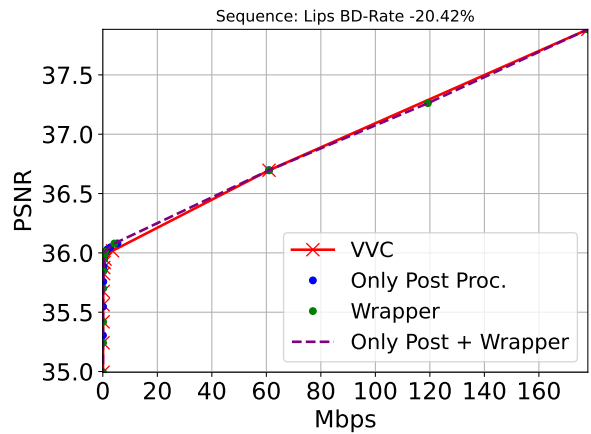

(a) Small range  (b) Large range

10

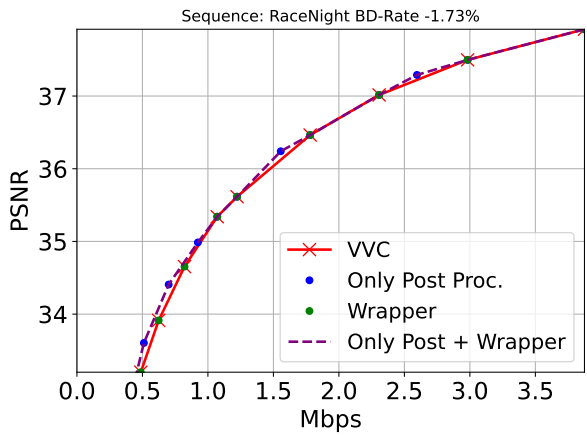

(a) Small range

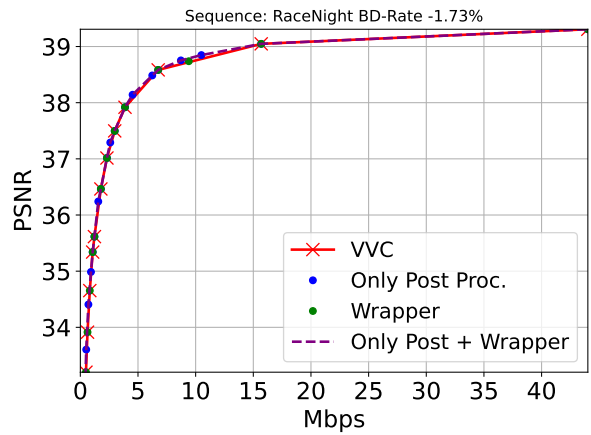

(b) Large range

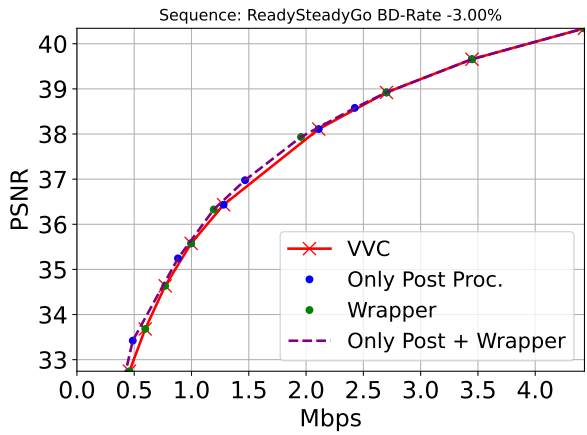

(a) Small range

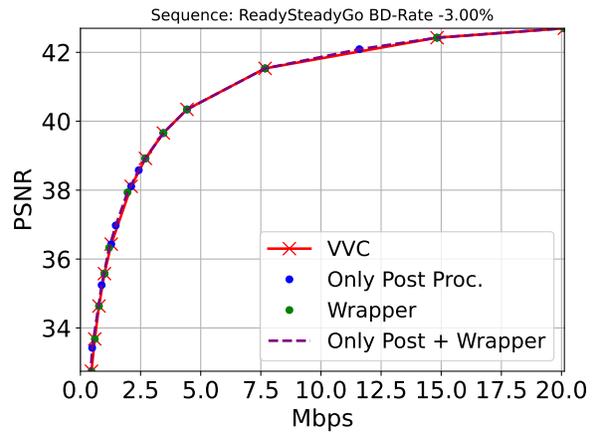

(b) Large range

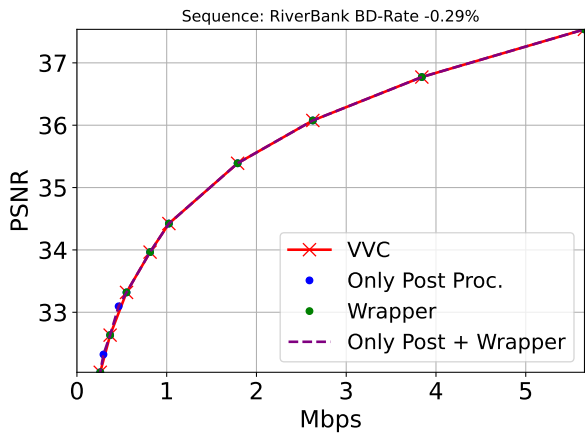

(a) Small range

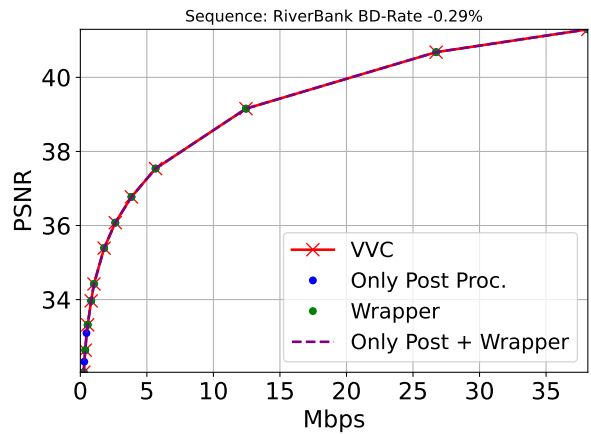

(b) Large range

11

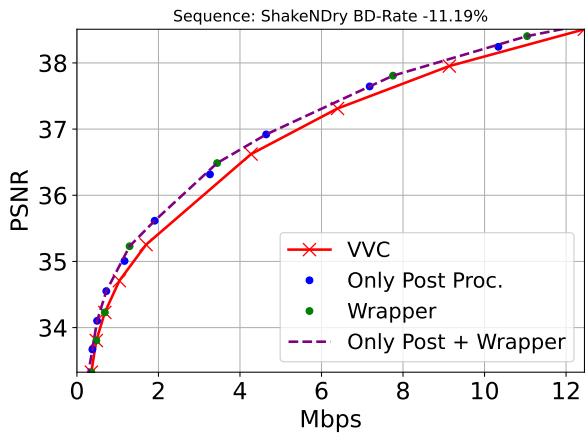

(a) Small range

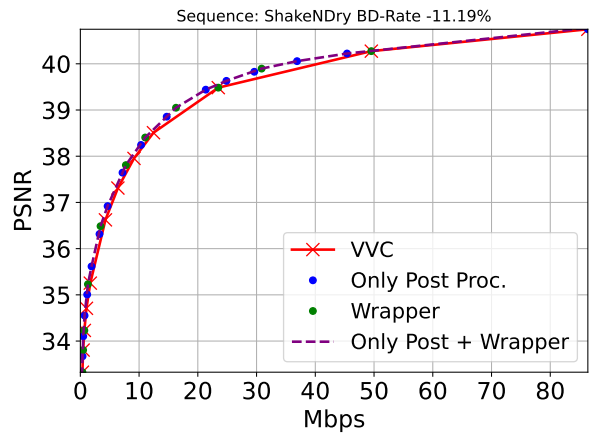

(b) Large range

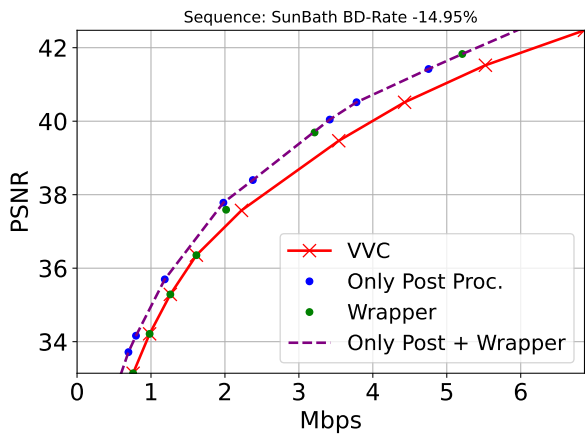

(a) Small range

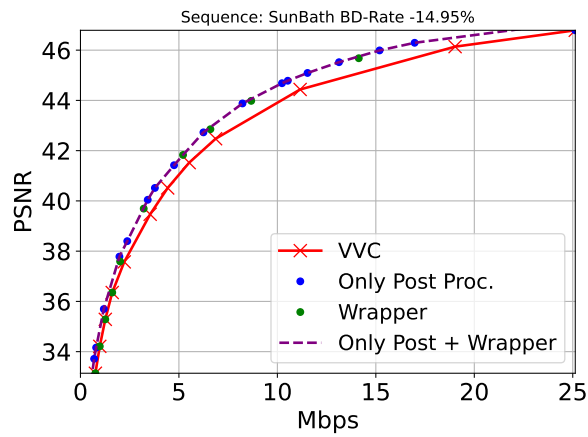

(b) Large range

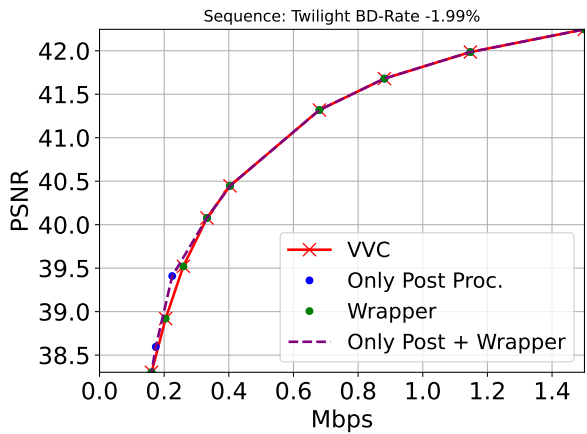

(a) Small range

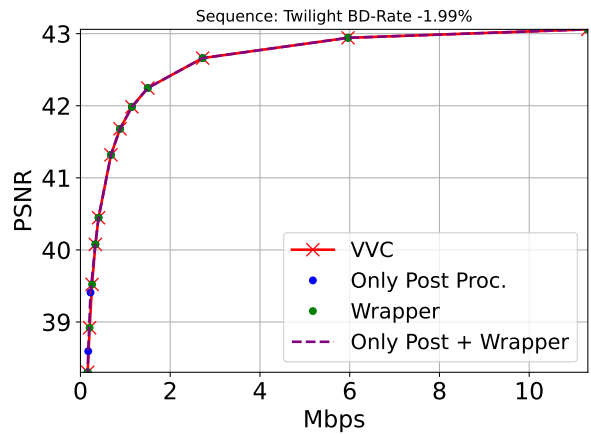

(b) Large range

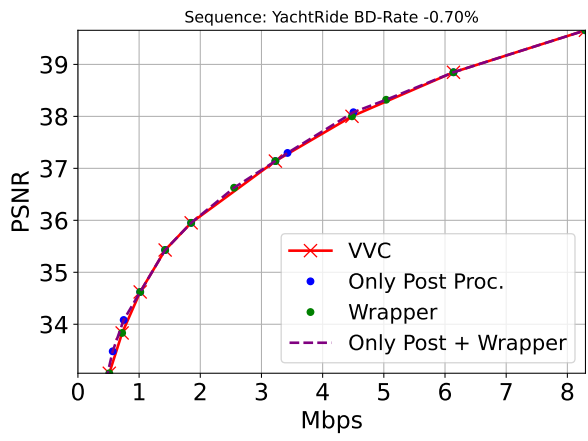 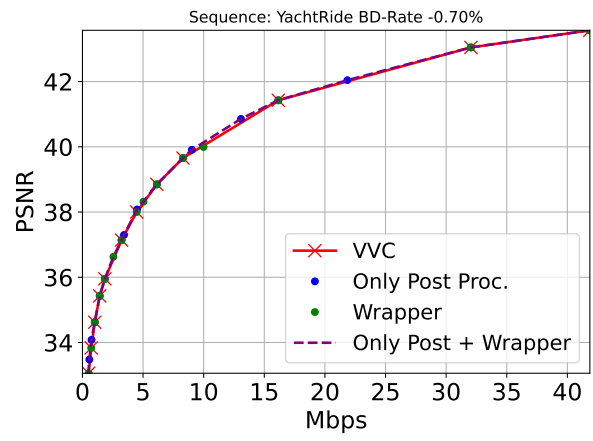

(a) Small range

(b) Large range

# PART III
## Rate-Distortion Curves on AOM CTC Class A1 with HEVC

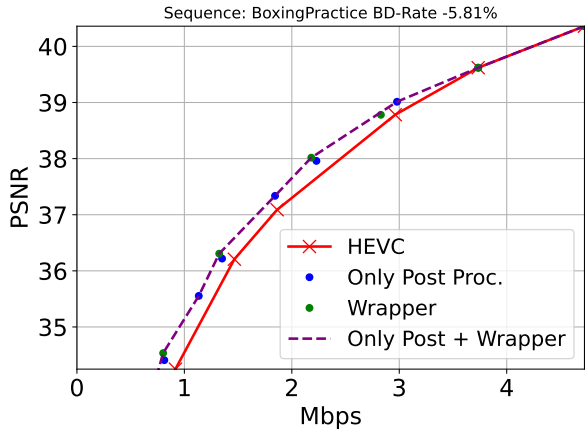

(a) Small range

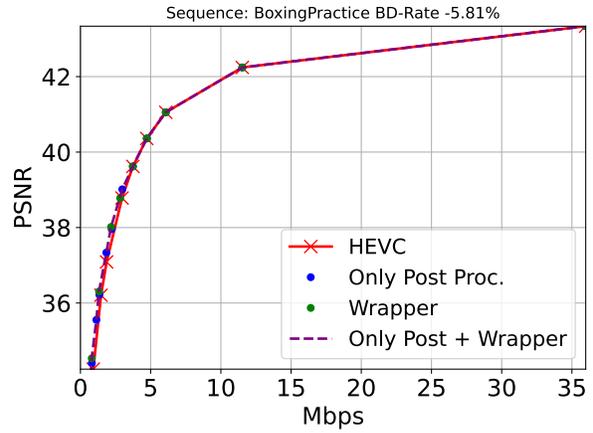

(b) Large range

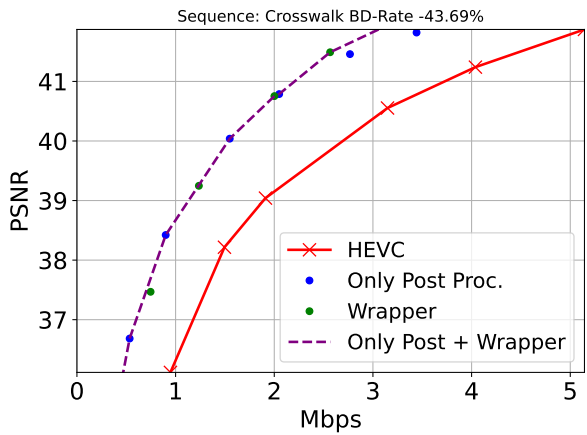

(a) Small range

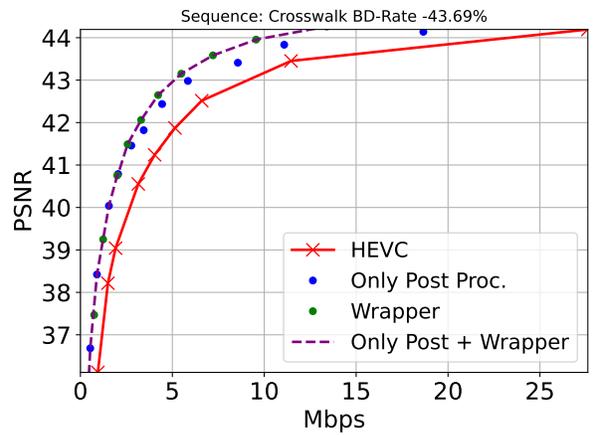

(b) Large range

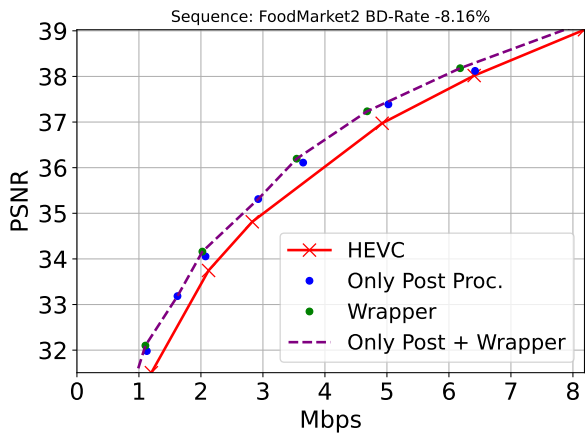

(a) Small range

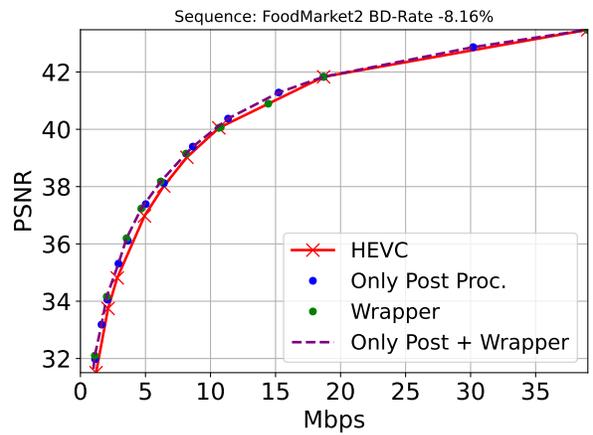

(b) Large range

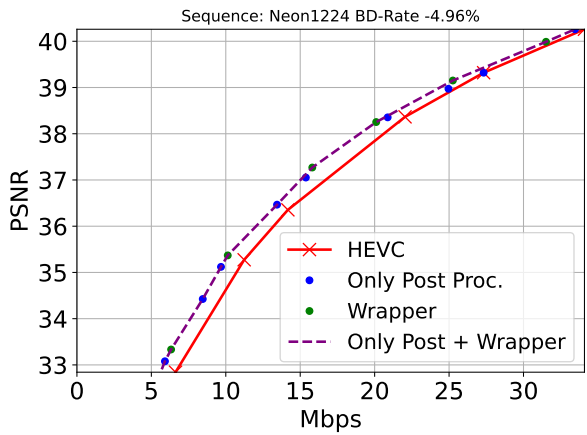
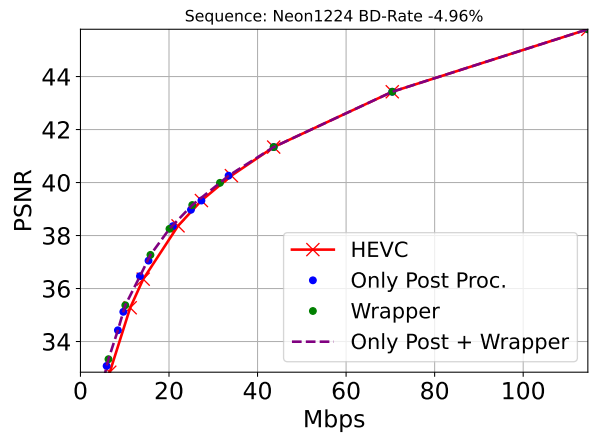

(a) Small range

(b) Large range

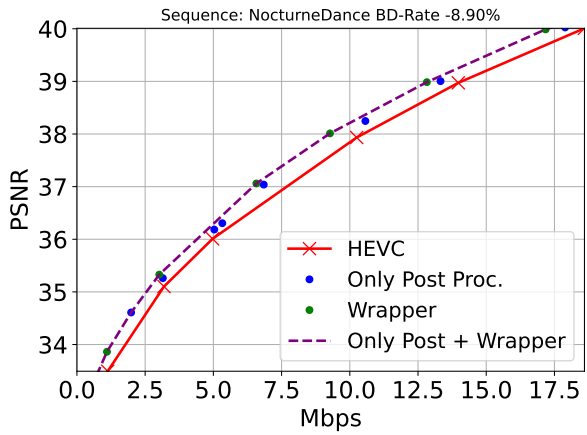
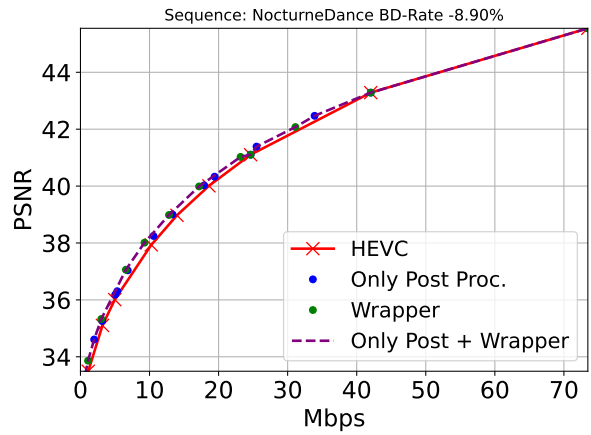

(a) Small range

(b) Large range

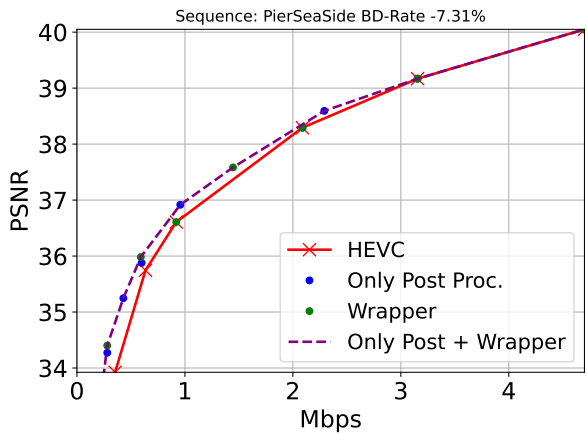
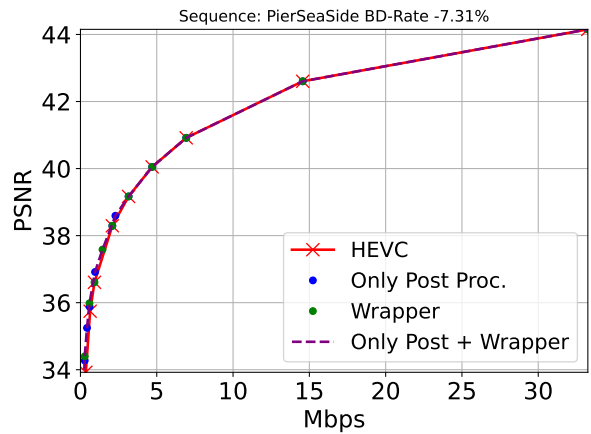

(a) Small range

(b) Large range

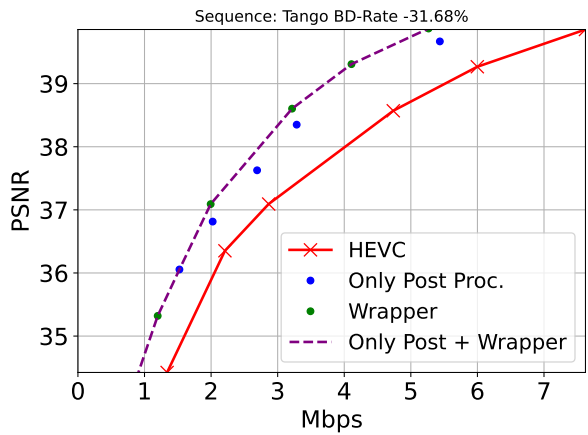
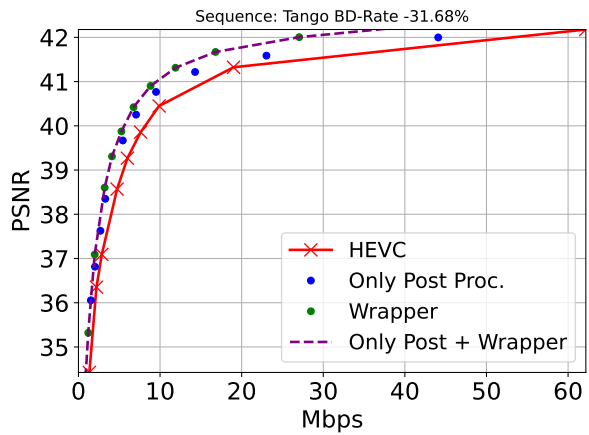

(a) Small range

(b) Large range

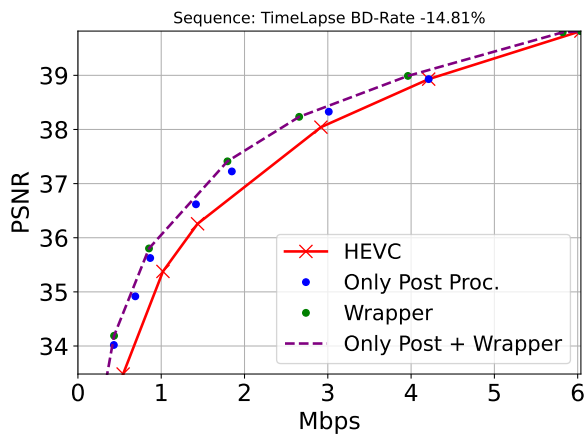
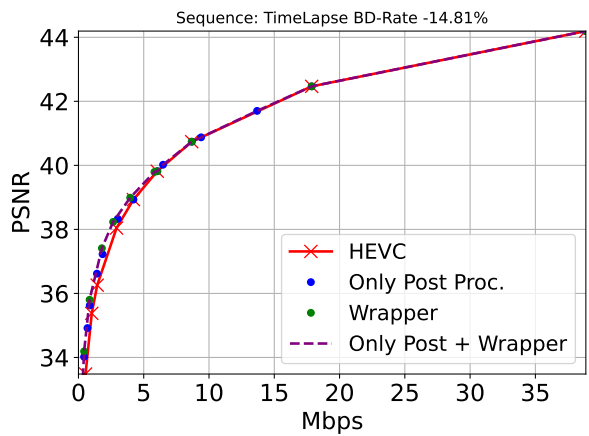

(a) Small range

(b) Large range

# PART IV
## Rate-Distortion Curves on AOM CTC Class A1 with VVC

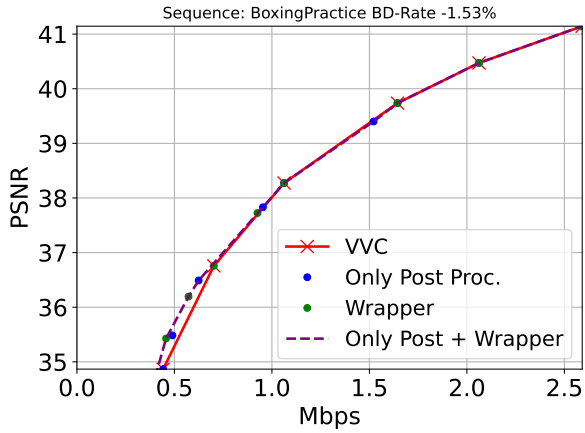

(a) Small range

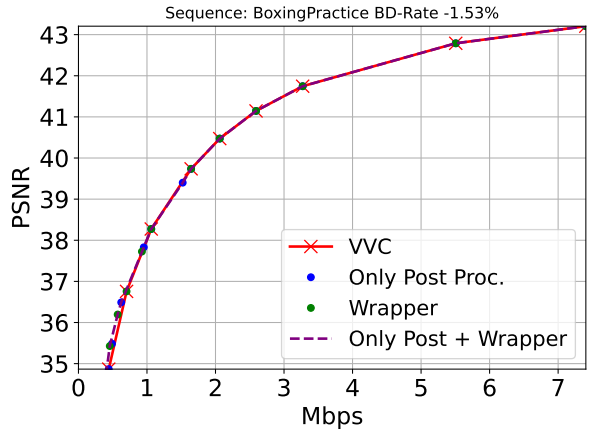

(b) Large range

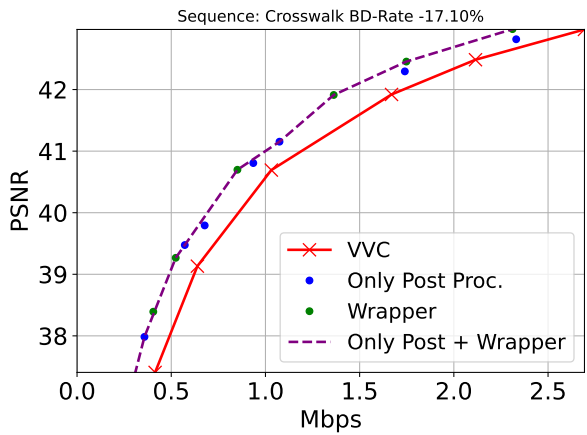

(a) Small range

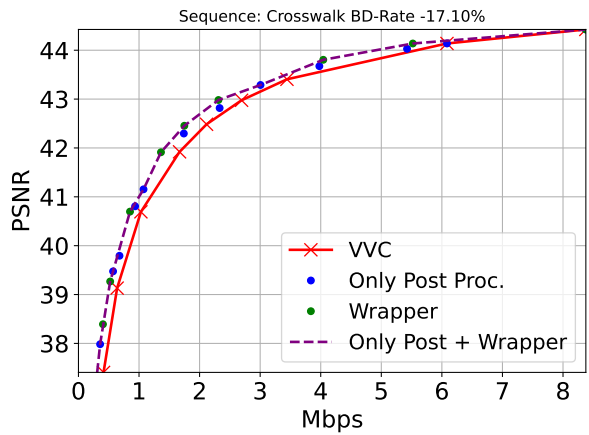

(b) Large range

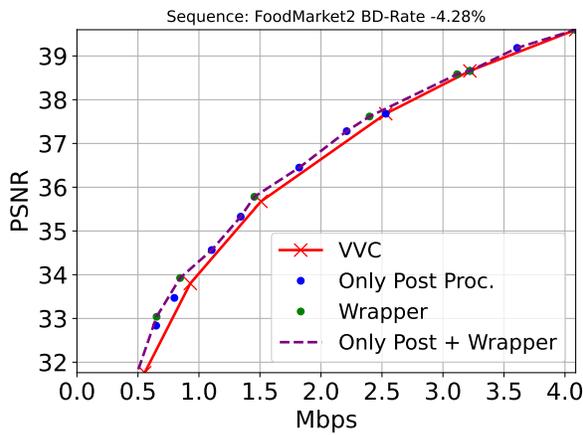

(a) Small range

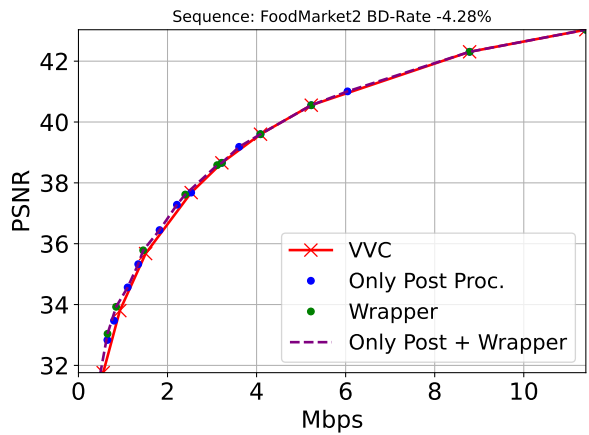

(b) Large range

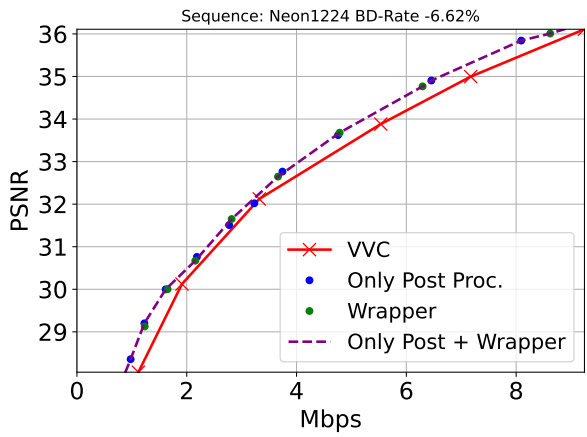 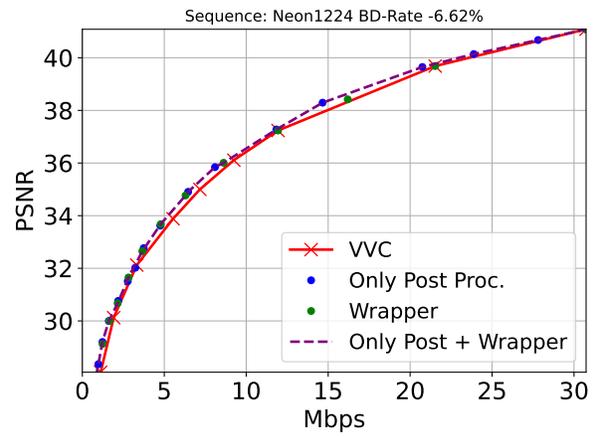

(a) Small range

(b) Large range

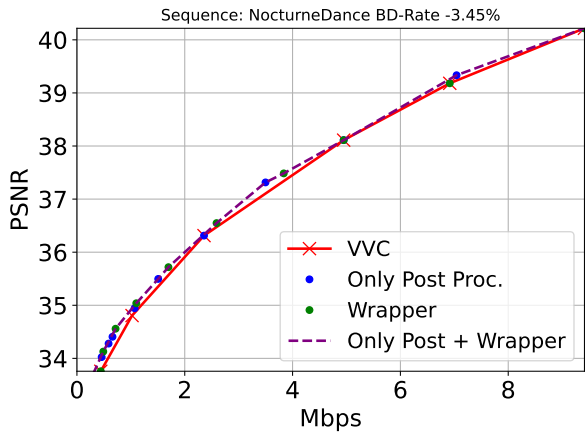 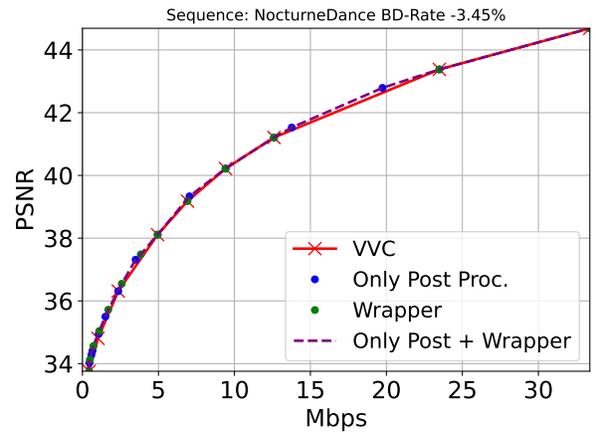

(a) Small range

(b) Large range

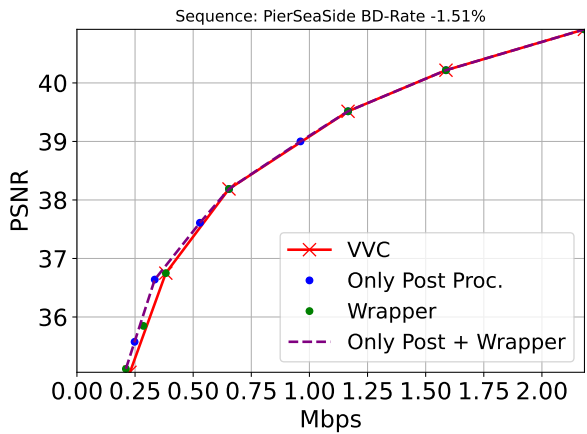 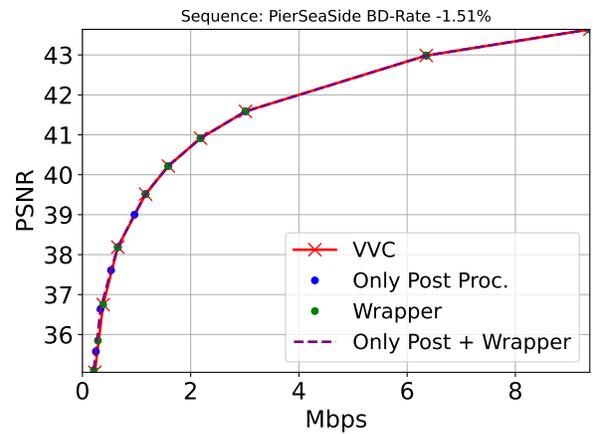

(a) Small range

(b) Large range

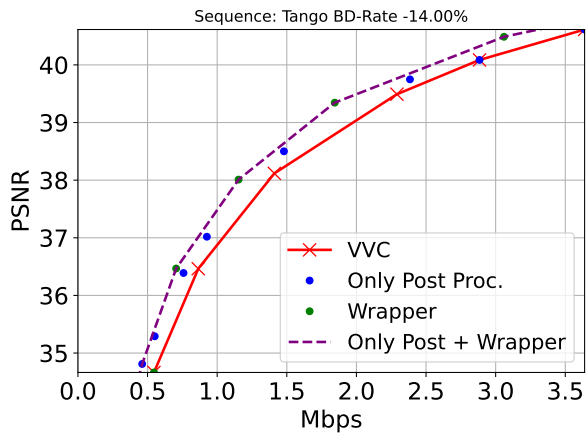
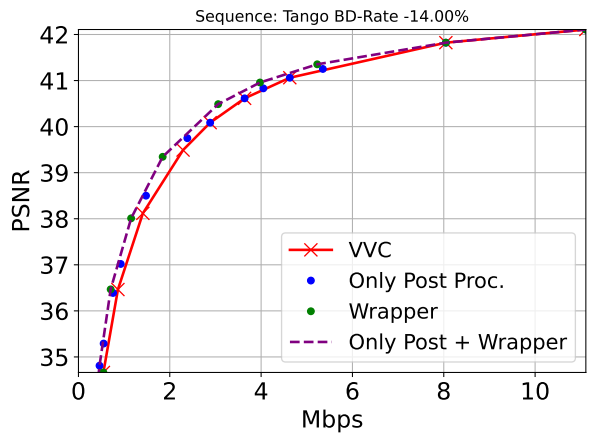

(a) Small range

(b) Large range

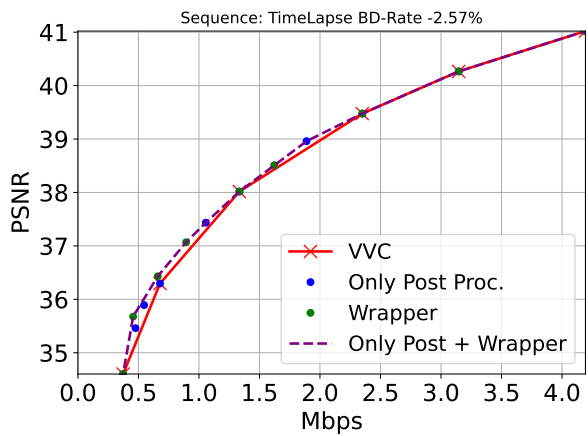
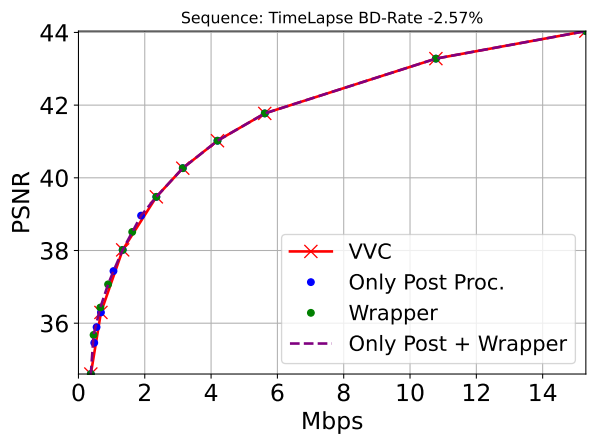

(a) Small range

(b) Large range

19



\end{document}